\documentclass[conference]{IEEEtran}
\IEEEoverridecommandlockouts
\usepackage{cite}
\usepackage{microtype}
\usepackage{graphicx}
\usepackage{subfigure}
\usepackage{booktabs} 
\usepackage{bigstrut}
\usepackage{amsmath}
\usepackage{amssymb}
\usepackage{bigstrut}
\usepackage{mathrsfs}
\usepackage{multirow}
\usepackage{url}

\usepackage{caption}

\usepackage{color}
\usepackage{balance}
\usepackage{dblfloatfix}
\usepackage{amsmath,amssymb,amsfonts}
\usepackage{algorithmic}
\usepackage{comment}

\usepackage{enumitem}

\usepackage{breqn}
\usepackage{dblfloatfix}
\usepackage{pgfplots}

\usepackage[framemethod=tikz]{mdframed}

\usepackage{textcomp}
\usepackage{xcolor}

\def\BibTeX{{\rm B\kern-.05em{\sc i\kern-.025em b}\kern-.08em
    T\kern-.1667em\lower.7ex\hbox{E}\kern-.125emX}}
\begin{document}

\title{ 
Face Morphing Attack Generation \& Detection:\\ A Comprehensive  Survey \\
	 }

\author{Sushma Venkatesh \quad   Raghavendra Ramachandra \quad Kiran Raja   \quad Christoph Busch\\
Norwegian University of Science and Technology (NTNU), Norway\\ 
E-mail:	\{\tt\small sushma.venkatesh;raghavendra.ramachandra;kiran.raja;christoph.busch\} @ntnu.no\\

}

\maketitle

\begin{abstract}
Face recognition has been successfully deployed in real-time applications including secure applications like border control. The vulnerability of Face Recognition System (FRS) to various kind of attacks (both direct and in-direct attacks) and face morphing attacks has received a great interest from the biometric community.  The goal of a morphing attack is to subvert the FRS at Automatic Border Control (ABC) gates by presenting the Electronic Machine Readable Travel Document (eMRTD) or e-passport that is obtained based on the morphed face image. Since the application process for the e-passport in the majority countries requires a passport photo to be presented by the applicant, a malicious actor and the accomplice can generate the morphed face image and to obtain the e-passport. An e-passport with a morphed face images can be used by both the malicious actor and the accomplice to cross the border as the morphed face image can be verified against both of them. This can result in a significant threat as a malicious actor can cross the border without revealing the track of his/her criminal background while the details of accomplice are recorded in the log of the access control system. This survey aims to present a systematic overview of the progress made in the area of face morphing in terms of both morph generation and morph detection. In this paper, we describe and illustrate various aspects of face morphing attacks, including different techniques for generating morphed face images but also the state-of-the-art regarding Morph Attack Detection (MAD) algorithms based on a stringent taxonomy and finally the availability of public databases, which allow to benchmark new MAD algorithms in a reproducible manner. The outcomes of competitions/benchmarking, vulnerability assessments and performance evaluation metrics are also provided in a comprehensive manner. Furthermore, we discuss the open challenges and potential future works that need to be addressed in this evolving field of biometrics.
\end{abstract}

\section{Introduction}
\label{sec:introduction}
Biometrics is a technique to recognize an individual based on unique biological (e.g., face, fingerprint, iris) or behavioral characteristics (e.g., gait, keystroke style) \cite{jain2007handbook} \cite{choras2019multimodal}. With the drastic improvement in the deep learning techniques, the biometrics-based person identification and verification has emerged as a popular technique that can be widely used for many secure access control applications. Ease to capture and suitability of the face biometric characteristics has further driven face recognition as a popular biometric modality in such applications. Face Recognition Systems (FRS) are widely deployed for various applications, especially in secure access control for person identification and verification purposes. Amongst several other applications like healthcare, law enforcement, e-commerce (banking), one of the most relevant applications is the border control process where the face characteristic of a traveler is compared with a reference in a passport or visa database, in order to verify the claimed identity. 

 \begin{figure*}[htp]
 \centering
\includegraphics[width=0.9\linewidth]{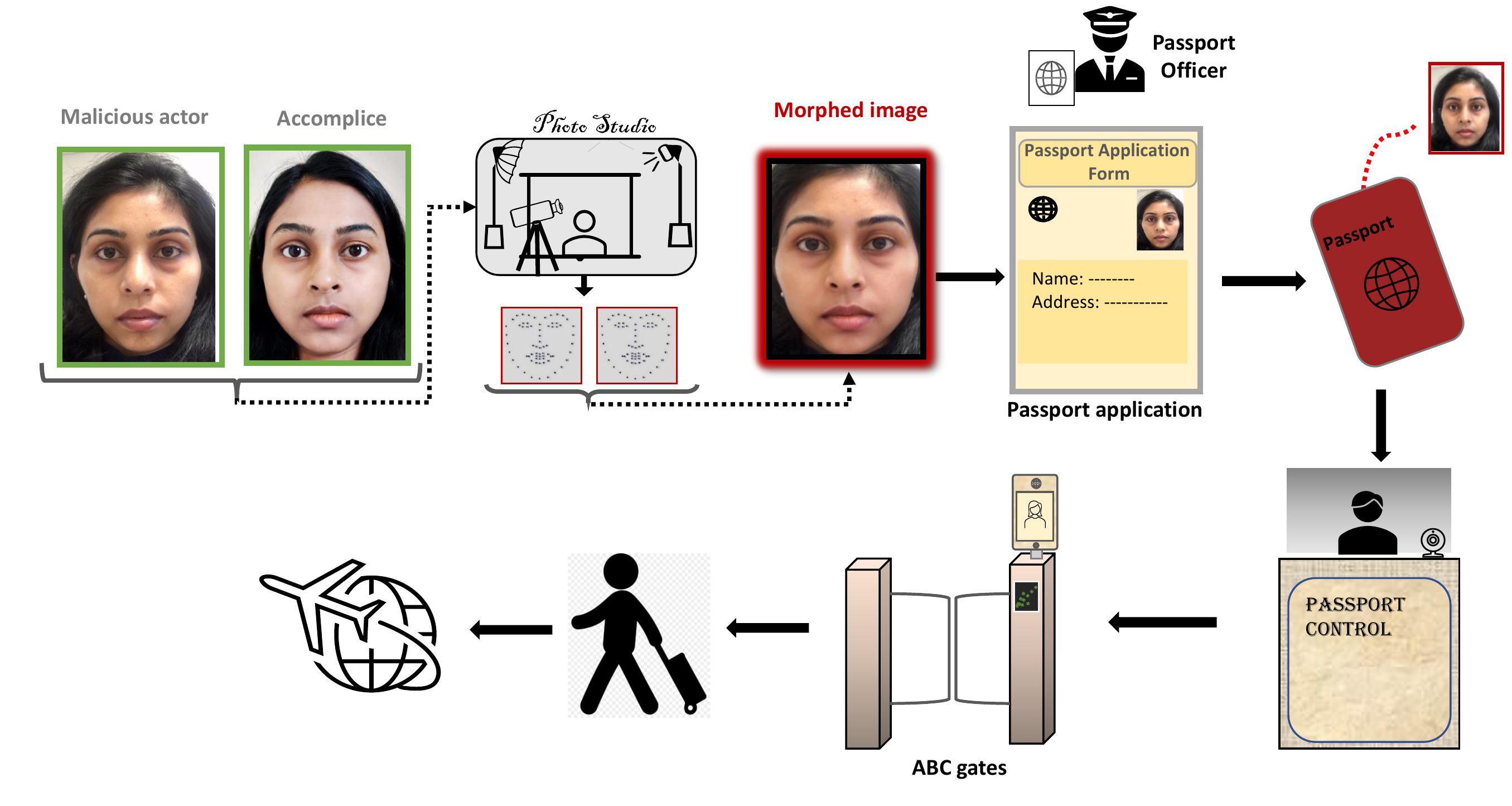}
\captionof{figure}{An example scenario illustrating the vulnerability of FRS from morphed images in border control scenario.}
\label{fig:Morph_vulnerability_ex_fig}
\end{figure*}

Though FRS effectively distinguishes an individual from other subjects, FRS's risk of being attacked to mislead/conceal the actual identity is a major concern. As with all the applications, the FRS is also prone to various attacks such as presentation attack with a goal to subvert the FRS by presenting an artefact\cite{PADsurvey_2017} where various types of attacks like electronic display attack, print attack, replay attack and 3D face mask attacks can be used \cite{PADsurvey_2017} \cite{Galbally_PAD} \cite{jia2020survey} \cite{anjos_photoattack} \cite{Spoof_3Dmask} \cite{replay_attack} \cite{Agarwal_2017_CVPR_Workshops} \cite{LiveFace_SpoofFace} \cite{evans2015guest} \cite{chingovska2014biometrics}. Besides these attacks, a morphing attack has also emerged in the recent past as severe threat to the enrolment process that undermines the FRS capabilities \cite{Ferrara2014} successfully. Face morphing is defined as ``a seamless transition of a facial image transforming a facial image into another" \cite{Morphing_ScienceDaily} in the context of biometrics, two or multiple facial images can be combined to resemble the contributing subjects. Morphing attacks raise a major concern as the morphed image represents the facial characteristics of both to the morphing process contributing individuals (for instance, an accomplice and a malicious actor).  Eventually, the resulting morphed facial image can successfully be verified with probe images from both contributing subjects making it practically usable for various malicious actions. Therefore, this attacks breaks the rule of single ownership, for instance, as for identification documents like a passport or electronic Machine Readable Travel Document (eMRTD) \cite{Passport_wiki} the unique link with the data subject for which the document was issued. The facial image stored in the eMRTD or passport is compared with the person claiming the identity document ownership while crossing the border. If the enrolled facial image reaches a match with the live image, the data subject can cross the border control. Thus an individual with malicious intention can exploit the face morphing attack and get illegal access. Hence a malicious person can easily cross a border using eMRTD or passport if he/she has contributed to the morphed image, which was used in the passport application process. 

Figure \ref{fig:Morph_vulnerability_ex_fig} illustrates an example scenario in the border control where the facial image of a malicious person is morphed with a look-alike accomplice. As there are several morphing software available freely, even a non-technical person can perform morphing with ease. The accomplice can submit the generated morphed image for passport enrolment at the passport issuance office. As the morphed image's facial features resemble the applicant's face, the passport officer approves the application. Eventually, a malicious person can successfully use the genuine passport allowing him/her for all foreseeable purposes (e.g., crossing of a border control).

In most countries, the applicant will submit the printed facial image to the passport office, leaving a possibility to provide a morphed image after print and scan. However, some countries like New Zealand, Estonia and Ireland also accept a digital facial image for passport renewal \cite{nzvisa2019portal}. Hence an applicant can submit a digital facial image to the web portal. This practice further raises a severe concern as there is no trusted supervision while uploading the digital facial image and opens the possibility of uploading a morphed image. Besides, the B1/B2 visa application for the United States also allows the applicant to upload a digital facial image in the web portal  \cite{UStravelVisa}.  An applicant can use this opportunity to upload a morphed image with the intent to perform illegal activity.

All such vulnerabilities of FRS have made morphing research very crucial in the recent years to avoid probable security lapse.
Thus, several research projects have been funded by the European Union and national research councils  (e.g. SWAN \cite{SWAN_project}, ANANAS \cite{BMBF_ANANAS_project}, SOTAMD \cite{SOTAMD_project} and iMARS \cite{iMars}) to focus extensively on creating Morph Attack Detection (MAD) algorithms.  Motivated by the momentum of the problem of morphing and its criticality, a dedicated conference has further been initiated by the Frontex, the European Border and Coast Guard Agency \cite{Frontex_Conf}  where a MAD interest group gathered to discuss the challenges and advancements of MAD techniques \cite{frontex2015best}. Further, the U.S. National Institute of Standards and Technology (NIST) is in parallel conducting testing of MAD technology within the framework of the Face Recognition Vendor Test (FRVT) under Part 4: MORPH - Performance of Automated Face Morph Detection\cite{NISTReport2020}. Both industrial and academic institutions are invited to submit their MAD algorithms to benchmark the accuracy \cite{NISTReport2020}.  In the similar lines, the University of Bologna within the SOTAMD project \cite{UBOFVCMorphing} has introduced a parallel face morphing evaluation platform to benchmark the performance of the MAD techniques on a sequestered dataset.

The rest of the survey is organized as follows:  Section \ref{sec:FMA} presents a brief introduction on face morphing attack, Section \ref{sec:morph_generation} discusses the face morph generation techniques. Section \ref{sec:database} describes the face morphing datasets that includes private and public,  Section \ref{sec:Human} discusses the human perception capabilities for detecting morphed face images, Section \ref{sec:MAD} presents various automatic morphing attack detection techniques, Section \ref{sec:performancemetrics} presents the performance metrics that are widely used to benchmark the performance of MAD methods as well as the vulnerability of the generated morphed image,  Section \ref{sec:Publiccompetitions} discuss the public evaluation and benchmarking of MAD,  Section \ref{sec:challenges} discusses the open challenges and potential future works and Section \ref{conclusion} draws the conclusion.

\section{Face Morphing Attack}
\label{sec:FMA}

 \begin{figure}[htp]
	\centering
	\includegraphics[width=1\linewidth]{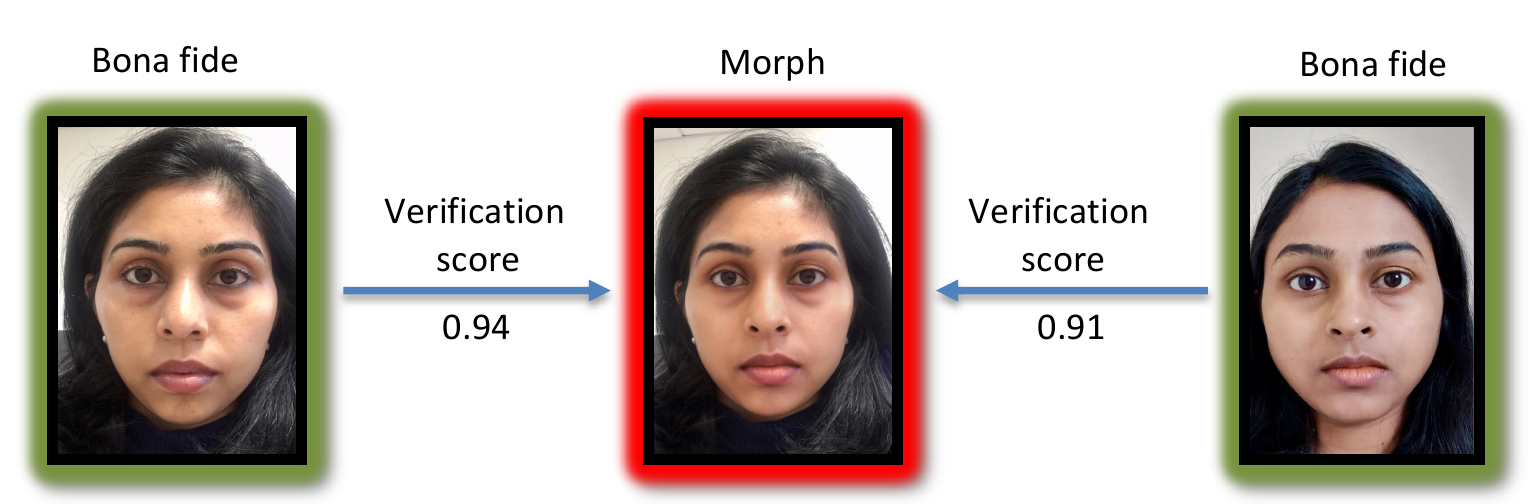}
	\caption{Impact of face morphing on FRS. As noted in the figure, the morphed image can be equally verified against both contributing subjects with a high similarity score from FRS ($1$ being high similarity).}
	\label{fig:Morph_intro_fig}
\end{figure}

The morphing process can be defined as a special effect that transforms one image into another image.
Figure \ref{fig:Morph_intro_fig} illustrates the facial morphing process, where two facial images are combined together to generate a single morphed image. Morphing can be easily achieved by using one of the numerous and freely available tools such as MorphThing \cite{MorphThing}, 3Dthis Face Morph \cite{3Dmorph}, Face Swap Online \cite{Faceswap}, Abrosoft FantaMorph \cite{FantaMorph}, FaceMorpher \cite{Facemorpher}, MagicMorph \cite{Magicmorph} 
The morphed image possesses near-identical features with both subjects contributing to generating morphs when subject pre-selection is applied (e.g look-alike)\cite{raghavendra2017face}.

Further, when processed with care, the morphed image does not possess many visible artefacts and thus, a human observer may fail to detect the image manipulation based on morphing. In practice, this leads to a situation where a passport officer may not be able to realize the morphing attack despite being an expert in facial comparison \cite{MAD_Human2019, Morph_PLUSONE}. It triggers a reasonable condition where a criminal with malicious intent can use a passport enrolled with a morphed image and cross the border without challenge. Figure \ref{fig:Morph_vulnerability_ex_fig} illustrates the vulnerability of Face Recognition Systems (FRS) when attacked with morphed image in the border control scenario. 

\begin{figure}[htp]
	\centering
	\includegraphics[width=1\linewidth]{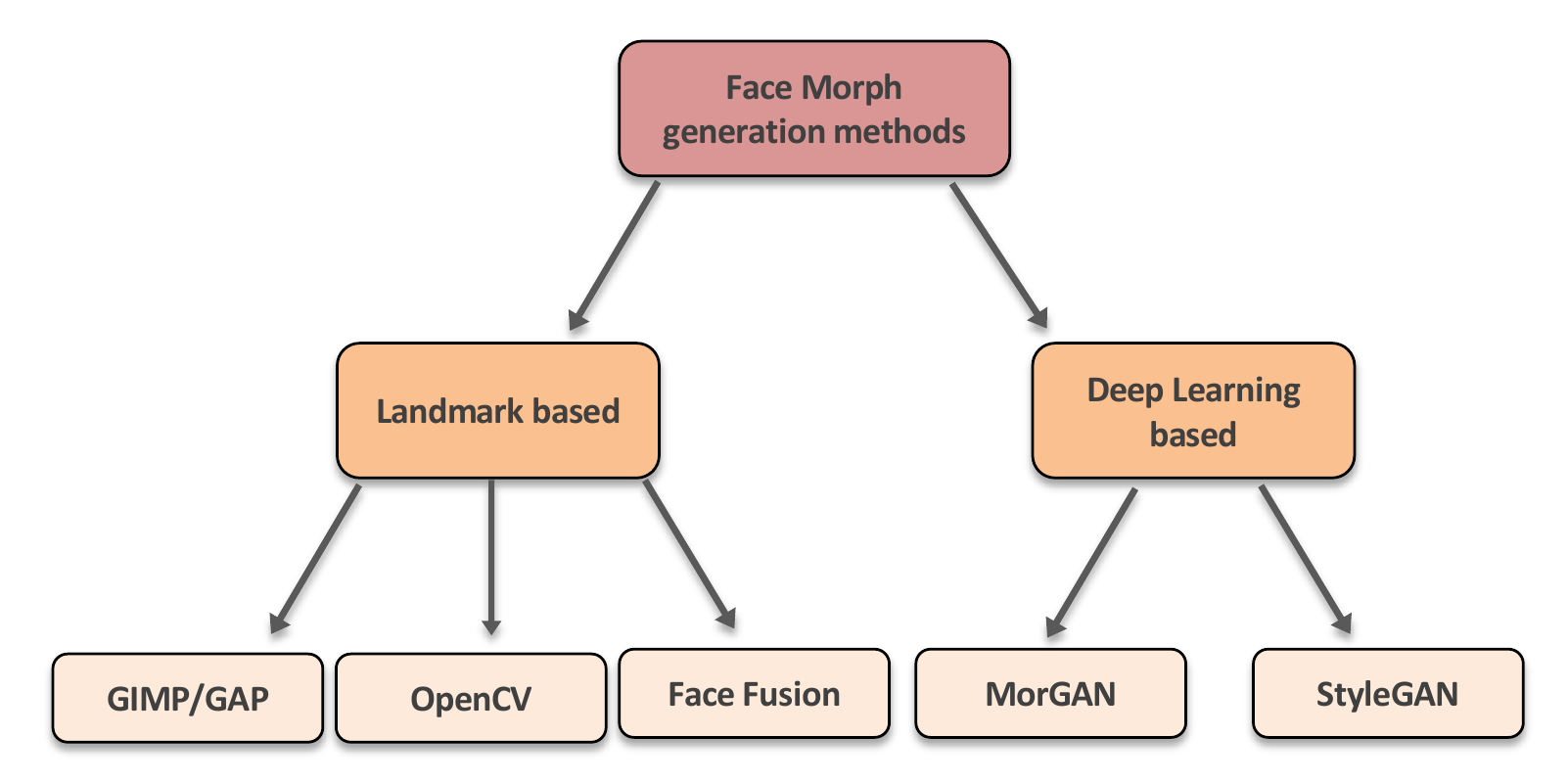}
	\vspace{-3mm}
	\caption{Taxonomy of face morph generation techniques}
	\label{fig:Morph_generation_taxonomy}
\end{figure}

\begin{table*}[htbp]
  \centering
  \caption{Face Morphing Generation Methods: Advantages and Limitations}
    \resizebox{0.9\linewidth}{!}{
      \begin{tabular}{|p{8.085em}|p{17.835em}|p{17.165em}|}
    \hline
    \multicolumn{1}{|c|}{\textbf{Face Morph}} & \multicolumn{1}{c|}{\textbf{Advantages}}  & \multicolumn{1}{c|}{\textbf{Limitations}} \\
    \multicolumn{1}{|c|}{\textbf{Generation Method}} &  & \bigstrut\\
    \hline
    Facial Landmarks based & - Availability of open source tools. \newline{} - Generate high quality morphing images.\newline{} - Successfully deceive the COTS FRS.\newline{}-Easy and seamless generation of \newline{} morphed images by automatic process.  & - Requires manual interventions to assure  \newline{} high quality face morphing generation.
\newline{} - Need post-processing to reduce \newline{} ghosting effect and double edges. \newline{} - Data subject selection is \newline{} crucial to deceive the COTS FRS.  \bigstrut\\
    \hline 
    Deep Learning Based & - No need of manual interventions. \newline{} - Seamless generation with acceptable image quality. \newline{} - Does not show double edges in the generated images. \newline{} - Reasonably successful in deceiving the COTS FRS. \newline{}  - Several open-source tools. & - Requires complex learning procedure. \newline{} - Not always generate high-quality morphed images. \newline{} - Highly prone to geometric distortions. \newline{} - Requires careful pre-selection on data \newline{} subjects based on age, gender and ethnicity.\\
    \hline
    \end{tabular}%
}
  \label{tab:limitations-comparison}%
\end{table*}%

\section{Face Morph Attack Generation}
\label{sec:morph_generation}
Face morphing was widely used for more than a decade, especially in the video animation industry \cite{Morph_entertainment}, but the attack potential on the FRS was noted in the recent past \cite{Ferrara2014}. Morphs can be generated using various techniques from simple image warping to recent Generative Adversarial Networks (GAN)  \cite{ImgMorph_Survey1998, arad-ImageWarping-CVGIP-1994, Lee-ImgMetamorphosisScatteredFeatureConstraints-Journal-1996, blanz-MorphableModel-Siggraph-1999, Zanella-AutomaticFaceMorph-Book-2009, 2013StudyLandmark, Liao-AutomatingFaceMorphing-ACMTrans-2014, ImageMorphAlgo_Survey2015, MIPGAN}. The most widely used morph generation is based on the landmark-based technique \cite{GIMP:2014, Raghavendra2016, ferrara_decoupling_2019, PRNU_TBIOM2019} where morphing is carried out by combining the images with respect to corresponding landmarks. Recent works eliminate the constraints of landmarks by simply relying on deep network architectures \cite{MorGAN, MIPGAN}. Figure \ref{fig:Morph_generation_taxonomy} shows the taxonomy of the face morphing generation that indicates the broad classification of the available techniques as (a) Landmark based and (b) Deep Learning-based approaches.


\begin{figure}[htbp]
	\centering
	\includegraphics[width=1\linewidth]{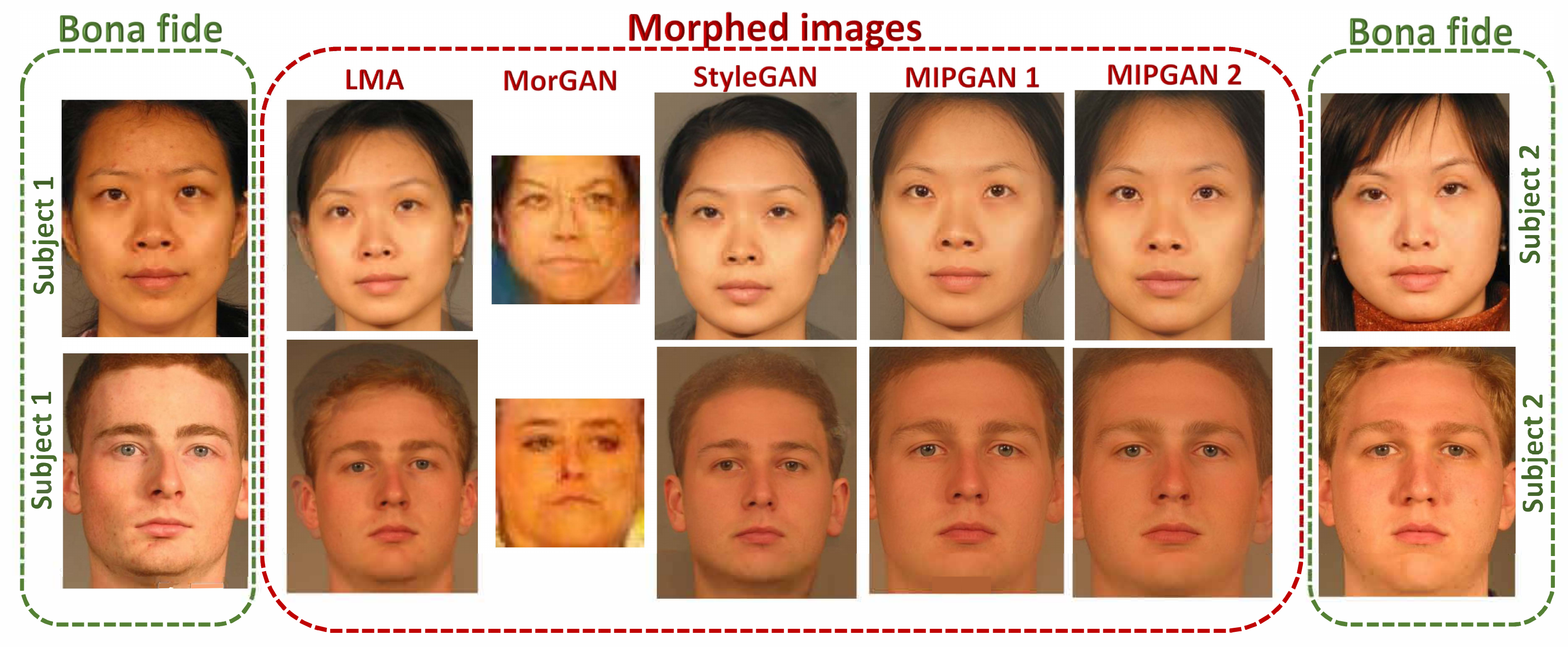}
	\vspace{-3mm}
	\caption{Illustration of face morphing image generated using different methods}
	\label{fig:Morph_comparison_images}
\end{figure}

\subsection{Landmark Based Morph Generation}
Landmark-based morph generation works on obtaining the landmark points on the face region, e.g., nose, eye, mouth region. These landmark points obtained from both faces are warped by moving the pixels to different more averaging positions. Different procedures for warping exist, including Free Form Deformation (FFD) \cite{lee1995ImgMetamorph, Warp1992feature}, Deformation by moving least squares  \cite{ImageDeformation-Schaefer2006}, deformation based on mass-spring \cite{ImgMorph-MassSpring-choi2011},  Bayesian framework based morphing \cite{Interpolation-morping-Bichsel1996}. Ruprecht et al.,  \cite{imageWarp-ruprecht1995} proposed to perform warping by moving the pixel points of both the contributory subjects to the nearest landmark point.  Delaunay triangulation was later proposed where the pixels of both the contributing facial images are distorted and moved to different directions to generate triangles  \cite{visapp17_Morph, FRusingMOrphwu2011, IWBF2017_StirTrace, seibold2017ONLYVGGdetection, Scher2017, FeatureDifference}. 
Images (that are to be morphed) are blended by considering the blending factors or the morphing factor. Face morphing applications employ a morphing factor of $0.5$ to generate good quality and useful morphs that can resemble both contributing subjects equally and to make the COTS FRS vulnerable \cite{Raghavendra2016, raghavendra2017face, raghavendra2017transferable}. As the morphing process replaces the pixel positions, there may be some misaligned pixels that contribute to noise generating artefacts and ghost-like images making the images unrealistic in appearance (i.e., easy to detect by the human observer). Hence, certain post-processing steps such as image smoothing, image sharpening, edge correction, histogram equalization,  manual retouching, and image enhancement improve the brightness and contrast that can reduce/minimise the artefacts generated during the morphing process are usually performed \cite{seibold2017ONLYVGGdetection, FAceSwap-Bitouk2008, HairInterpolation-Weng2013}. 

The face morph generation using open source resources like GIMP/GAP and OpenCV also relies upon landmarks. While open-source software based on GIMP/GAP and OpenCV can generate morphs, a significant effort must be dedicated to post-process the generated images, in order to eliminate the artefacts. Several commercial solutions like Face Fusion \cite{scherhag2019face}, FantaMorph \cite{FantaMorph} can also be used to generate large scale morphed images with reasonable efforts for post-processing. The reader is further referred to Scherhag et al.,\cite{scherhag2019face} where all the publicly available morphing tools (both open source and commercial) are listed.

\subsection{Deep Learning Based Morph Generation}
Recent improvements in the deep learning-based techniques have given rise to morph generation approaches based on Generative Adversarial Network (GAN) \cite{MorGAN} \cite{StyleGAN_morph_2020}. In general, GAN based methods synthesize morphed images that are generated by sampling two facial images in the latent space of the deep learning network. MorGAN architecture for morph generation basically employs a generator that constitutes encoders, decoders and a discriminator. The generator is trained to generate the images with the dimension $64\times64$ pixels.  Another recent approach based on StyleGAN architecture \cite{image2style_GAN, StyleGAN_morph_2020} has improved the morph generation process both by increasing the spatial size to 1024$\times$1024 and face quality. The pre-trained StyleGAN achieves this by embedding the images in the intermediate latent space.  The use of identity priors to enable a high quality morphed face generation is further proposed in \cite{MIPGAN} and illustrates the increased threats to FRS amongst GAN based morphs. Figure \ref{fig:Morph_comparison_images} provides sample facial morphs generated using the landmark-based technique, MorGAN and StyleGAN based methods.  It can be noted from Figure \ref{fig:Morph_comparison_images}, that the deep learning-based approaches, especially with MIPGAN-I and MIPGAN-II, indicate a superior quality of the morphed face image and also compared to the landmark based morph face generation. 

\section{Databases on Morphing Attack Detection}
\label{sec:database}

Given various kinds of attack generation mechanisms and relevant attack potential determination metric, many datasets are generated ranging from public to sequestered dataset with various attack strengths. This section summarizes the different face morph databases that are used in existing works. A summary of different dataset is provided in Table \ref{tab:MorphDB} from the existing works that are typically used to benchmark both vulnerability of FRS and also the performance of MAD techniques.  

The first face morph database was introduced by Ferrara et al., \cite{Ferrara2014} where the authors have employed landmark based face morph generation using GIMP GAP tools. This dataset has a small set of digital images consisting of only 14 morphed images generated from 8 bona fide subjects that includes both male and female participants. The morphed images in this database are only in digital format and the database is not available publicly. This dataset was further extended  by Ferrara et al. \cite{Ferrara2016} using the landmarks and GIMP GAP tools. The extended dataset consists of around 80 morphed face images with 10 male and 9 female participants and the database is in digital form and it is not publicly available. 

The first large database with different ethnicity (Caucasian, Asian, European, American, Latin American, Middle Eastern) was introduced by Raghavendra et al., \cite{Raghavendra2016}, that employs facial landmarks and  GIMP GAP morph generation technique using the GNU image manipulation tool. This database consists of 450  morphed face images generated using 110 subjects of different ethnicity background. This database contains only digital images and was not made public.

\begin{table*}[htbp]
	\centering
	\caption{Public and private morph face image database}
	\resizebox{0.9\textwidth}{!}{
		\begin{tabular}{|p{10.835em}|p{6.765em}|p{6.235em}|p{6.235em}|p{12.065em}|p{6.235em}|}
			\hline
			\textbf{Reference} & \textbf{Morph Generation Type} & \textbf{Morph Generation Method} & \textbf{Digital/print-scan} & \textbf{Bona fide \& Morph} & \textbf{Public/Private} \bigstrut\\
			\hline
			Ferrara et al \cite{Ferrara2014} & Landmark based & GIMP GAP & Digital &   Morph: 14 & Private \bigstrut\\
			\hline
			Ferrara et al \cite{Ferrara2016} & Landmark based &GIMP GAP & Digital  &   Morph: 80 & Private \bigstrut\\
			
			\hline
			Raghavendra et al \cite{Raghavendra2016} & Landmark based  & GIMP GAP & Digital  &   Morph: 450 & Private \bigstrut\\
			\hline
			Makrushin et al \cite{visapp17_Morph} &  Landmark based & Automatic generation (dlib landmark) & Digital &  Complete morph: 1326, \newline{} Splicing morph: 2614  & Private \bigstrut\\
			\hline
			Scherhag et al \cite{Scher2017} & Landmark based &GIMP GAP& Digital and Print-Scan & Bona fide: 462  \newline{} Morph: 231 & Private \bigstrut\\
			\hline
			Raghavendra et al \cite{raghavendra2017face} & Landmark based &GIMP GAP &Digital and Print-Scan & Bona fide: 1000 \newline{} Morph: 1423+1423 & Private \bigstrut\\
			\hline
			Raghavendra et al \cite{raghavendra2017transferable} & Landmark based &GIMP GAP &Digital and Print-Scan &   Morph: 362 & Private\bigstrut\\
			\hline
			Gomez-Barrero et al \cite{Gomez_2017} & - & -     & Digital &   Morph: 840 & Private\bigstrut\\
			\hline
			Dunstone \cite{Biometix_MorphDatabase} & - & -     & Digital & Morph: 1082 & Public\bigstrut\\
			\hline
			Ferrara et al \cite{ferrara2018face}& Landmark based & Sqirlz morph & Digital and Print-Scan &   Morph: 100 & Private \bigstrut\\
			\hline
			Damer et al \cite{MorGAN} & GAN based &GAN & Digital &  Morph: 1000 & Private \bigstrut\\
			\hline
			Raghavendra et al \cite{RagCVIP2018} & Landmark based &GIMP GAP & Digital and Print/Scan & Bona fide: 1272 \newline{} Morph: 2518 & Private \bigstrut\\
			\hline
			Scherhag et al \cite{PRNU_TBIOM2019} & Landmark based & OpenCV, FaceFusion, Face Morpher & Digital and Print/Scan & Bona fide: 984+984+529 \newline{} Morph: 964+964+529 & Private \bigstrut\\
			\hline
			Ferrara et al \cite{ferrara_decoupling_2019} & Landmark based & Triangulation &Digital &   Morph: 560 & Private \bigstrut\\
			\hline
			Scherhag et al \cite{scherhag2020deep} & Landmark based  &OpenCV, FaceFusion, Face Morpher, UBO morpher &Digital and Print/Scan & Bona fide: 791+3298 \newline{} Morph: 791+3246 & Private\bigstrut\\
			\hline
			Singh et al \cite{Jag_ABC_gate_2019}  & Landmark based  &OpenCV &Digital and Print/Scan &   Morph: 588 & Private \bigstrut\\
			\hline 
			Venkatesh et al \cite{Venkatesh_2020_IJCB}  & Landmark based  & UBO morpher &Digital &   Morph: 10538+3767 & Private \bigstrut\\
			\hline 
			Venkatesh et al \cite{StyleGAN_morph_2020}& GAN based & StyleGAN & Digital & Bonafide: 1270 \newline{} Morph: 2500 & Private \bigstrut\\
			\hline
			Raja et al \cite{Raja2020MorphingAD}& Landmark based & UBO morpher  & Digital and Print/Scan & Bonafide: 300+1096 \newline{} Morph: 2045+3073 & Sequestered \bigstrut\\
			\hline
			NIST-FRVT-MORPH et al \cite{MeiNGAN-morph-FRVT-2020}& Landmark based &  Automatic generation  & Digital and Print/Scan &  Low quality morph: 1183 \newline{} Automated morph: 39113 \newline{} High quality morph: 492  & Sequestered \bigstrut\\
			\hline
		\end{tabular}%
	}
	\label{tab:MorphDB}%
\end{table*}%

Makrushin et al., \cite{visapp17_Morph} employed automatic morph generation tools to generate high quality morph images.  They have employed the triangulation method based on 68 facial landmarks extracted using dlib library \cite{dlib}.  Two different morph generation techniques namely; complete morph (consists of facial geometry of both the facial images) and splicing morph (pixels representing the face are clipped out from the input faces). 
Splicing morph is generated to overcome the pixel discontinuity caused while warping two images in complete morphs. This database consists of around 1326 complete morphs and 2614 splicing morphs generated  from 52 data subjects consisting of 17 female and 35 male. This database consists of face morph images in digital format only and it’s not made public. 

The first print/scan face morph database was presented by Scherhag et al.,\cite{Scher2017}. The authors have employed the landmark based GIMP GAP technique for morph generation. This database consists of 231 morphed images generated from 462 bona fide images. This database is private and contains digital and print-scan (or redigitized) images, for which HP Photosmart 5520 and Ricoh  MPC 6003 SP printers were employed.
 
Raghavendra et al., \cite{raghavendra2017face} later on introduced a new face morphing dataset consisting of both digital and print/scan images. The face morphs are generated using an automatic tool using OpenCV that is publicly available. This database generates morphed face images along with averaged face images and hence it has a set of $1423+1423$ morphed face images. Along with the database, Raghavendra et al., \cite{raghavendra2017face} also provided the evaluation protocol by defining independent sets for development, training and testing partition. The print-scan morphed face images are obtained by employing a Ricoh MPC 6003 SP printer. This database is private
This dataset was further extended to have 2518 morphed face images and  1273 bona fide images \cite{RagCVIP2018}.  

Gomez et al.,\cite{Gomez_2017} introduced a new  face morphing dataset that consists of 840 morphed face images generated from 210 subjects. This database is private and has only digital morphed face images. 
 
Ferrara et al.\cite{ferrara2018face}, \cite{Matteo_PrintScan_2019} introduced a face morphing database based on the Sqirlz morphing technique. This dataset has 100 morphed images in both digital and print/scan form. This database is not made public for research purposes.  Scherhag et al., \cite{PRNU_TBIOM2019} introduced a face morphing dataset that is generated using different morphing tools such as OpenCV, FaceFusion and FaceMorpher. This is a private database that consists of both digital and print/scan samples of morphed images and is composed of $964+964+529$ morphed face images generated from subjects contained in the FRGCv2 and FERET database. Another database by Scherhag et al.,\cite{scherhag2020deep} employs landmark based morph generation techniques that includes OpenCV, FaceMorphed, FaceFusion and the UBO morphing method. This database consists of around 791+3246 morphed face images from FERET and FRGCv2 database respectively. This private database consists of morphed face images in both digital and print/scan format. Another database by Ferrara et al.,\cite{ferrara_decoupling_2019}, employs the triangulation with dlib landmark method of morph generation. This is a private database that consists of 560 digital morphed face images. The only publicly available morph face dataset is introduced by Biometix \cite{Biometix_MorphDatabase}, which consists of 1082 morphed face images in digital form. However information on the involved morph image generation is not available. 


\begin{figure*}[htp]
	\centering
	\includegraphics[width=1\linewidth]{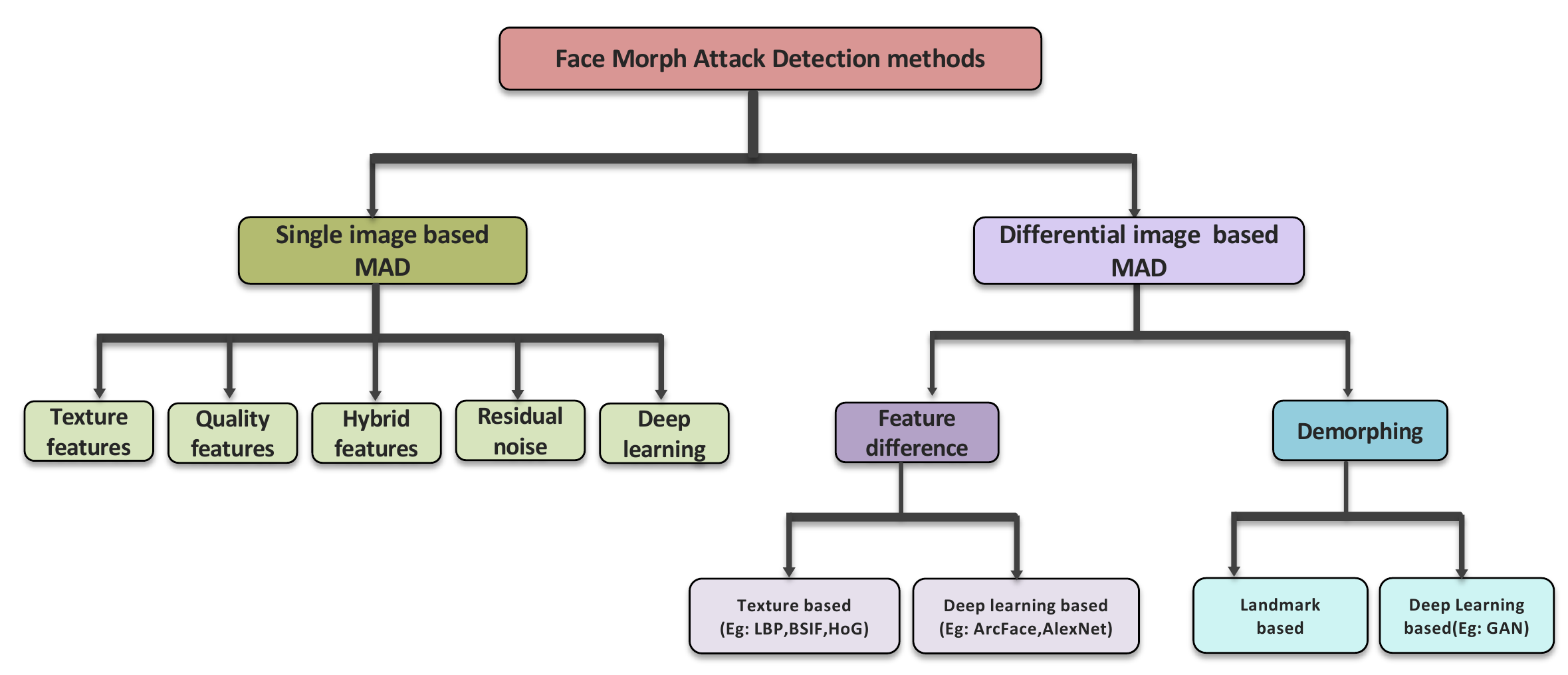}
	\vspace{-3mm}
	\caption{Taxonomy of Morph Attack Detection techniques}
	\label{fig:MAD_techniques}
\end{figure*}  

Singh et al \cite{Jag_ABC_gate_2019} provided another database that employs OpenCV based morph generation technique to generate the facial morphs. This is the first dataset introducing for probe images live-captures from ABC gates with different lighting conditions, which is relevant for differential morphing attack detection. This database consists of both digital and print-scan enrolment images generated using a EPSON XP-860 printer and scanner. This dataset consists of 90 morphed face images and it is not available for the public. 

Damer et al.,\cite{MorGAN} introduced the first face morphing database consisting of deep learning based morph images. The generated deep learning based database is compared with the landmark based morphs. The authors have employed 68 landmark points extracted from dlib for landmark based morph generation and GAN architecture for deep learning based morph generation. This database consists of 1000 morphed face images, however the GAN based morphs are of size 64$\times$64 which is not up to the ICAO standards. This database is private and has only digital morphed face data. Another database by Venkatesh et al.\cite{StyleGAN_morph_2020} employs deep learning based morph generation. The authors have employed StyleGAN network to generate synthetic morph images by mapping the input images into the latent space.  This database consists of 2500 morphed images generated using 1270 bona fide images. It has only digital morphed face images and it is not publicly available.

Venkatesh et al.\cite{Venkatesh_2020_IJCB} also introduced another database that consists of morphed face images under aging as a first of its kind. The authors have employed the UBO morphing method from the University of Bologna that employs dlib 68 landmark points for morph generation \cite{ferrara_decoupling_2019}. This database consists of 14305 (10538+3767) morphed face images with age spanning from 2 to 5 years. This database has morphed face images in digital form and it is not open for public. 

Raja et al. \cite{Raja2020MorphingAD} presents the sequestered Bologna-SOTAMD face morphing dataset used in the recent public competition and benchmarking, following the FVC-onGoing series \cite{FVCOngoingMorphingEval}. The dataset comprises images stemming from 150 data subjects collected in three different geographic locations with varying ethnicity, gender and age. The face morphing is carried out using six different techniques followed by automatic and manual post-processing to override the artefact results from the face morphing. The dataset also includes the printed and scanned versions with different printers, and the enrolment images are following the ICAO standard passport images. The probe images are taken from various ABC gates and gate emulations. The database consists of 5,748 morphed face images and 1,396 bona fide face images.

\subsection{Discussion}
Even-though there exist several morphing datasets, the majority of them are private due to data protection regulations and licensing conditions. Even for publicly available face databases that are used to create face morphing datasets the licensing conditions limit the redistribution of the generated morphed faces dataset and thereby most of the above discussed datasets are not openly available. For the time being the best way to compare new morphing detection methods with already published approaches is to submit the method to the two ongoing benchmarks, either the SOTAMD benchmark at the university Bologna, which was reported by Raja et al. \cite{Raja2020MorphingAD} or the U.S. NIST-FRVT-MORPH benchmark which is reported by Mei et al., \cite{MeiNGAN-morph-FRVT-2020}. Note in both cases a sequestered dataset is used.

\section{Human Perception and Morphed Face Detection}
\label{sec:Human}
The threat of morphing attacks is known for border crossing or ID management scenarios. Therefore, the success of a morphing attack depends on deceiving human observers, especially an ID expert or border guard.  The practical scenario for a border crossing includes those border guards, who compare the passport of traveller containing a photograph (printed from the data page and digital extracted from the chip) with a physical appearance of traveler. Thus, the border guard conclude based on the facial similarity of the traveler with the reference data in the passport for his final decision. Several studies in the literature indicated the effectiveness of morphed images in deceiving expert human observers \cite{Ferrara2016, Morph_PLUSONE, art29, RobertSon2018, MAD_Human2019, visapp17_Morph, MakrushinHuman, MakrushinHuam2, Jger2005PictureDO}. An early investigations on human perception analysis of morphed images was reported by J{\"a}ger et al.,  \cite{Jger2005PictureDO} where different experiments were performed to benchmark the ability of human observers to detect face morphing and its dependency to various parameters (i.e., different alpha/morphing-factors). While this was an interesting work, the human observers in the experiments were students not trained to compare human faces. Further, this work is based on only one single image without providing any reference image for the human observers. A similar analysis was provided by Kramer et al., \cite{MAD_Human2019} where a single image was provided before requesting a decision on morphing. Despite being different in the underlying benchmarking mechanism, both works have reported the difficulty in detecting morphed face images by human observers.

Investigating the impact of morphing on FRS and human observers simultaneously, Ferrara et al., \cite{Ferrara2014} studied the detection ability of human observers and correlated it to automatic FRS. Unlike the previous work, the human observers in the work by Ferrara et al., \cite{Ferrara2014} included both trained border guards and non-specialists who were asked to compare the morphed face image with a bona fide face image to make the decision. The analysis reported the challenge in detecting morphs even when the examiner, for instance a border guard, were trained. Robertson et al., \cite{Morph_PLUSONE} further presented the morph detection ability of humans by comparing the live face images to the morphed face images with and without rudimentary training.  The study reported improved performance of morph detection by human observers when provided with rudimentary training of detecting artefacts \cite{RobertSon2018}. Along a similar line, Kramer et al., \cite{MAD_Human2019} investigated the role of face image quality (of the achieved morphed image) on human perception and concluded that high quality morph images are more difficult to be detected by humans. 

A similar web-based experiment simulating border control was also presented by Makrushin et al.,\cite{MakrushinHuman} which studied  human perception analysis on both skilled and unskilled humans and further extended \cite{MakrushinHuam2} to have more unbiased and realistic images. In both cases, skilled humans (who have knowledge of morphed face images) show the best performance in detecting the morphed face image.  Summarizing the works on human perception analysis, it can be noted that both skilled and unskilled human observers drastically fail to detect a morphed face image. However, it is also noted that considerable training of human observers can improve the morphed face detection \cite{MakrushinHuam2, RobertSon2018}. 

\section{Face Morph Attack Detection Techniques}
\label{sec:MAD}
Noting the limitations of human observers, a number of automatic Morphing Attack Detection (MAD) approaches have been proposed in the recent past. In this section, we summarise the MAD techniques since the introduction of face morphing attack on FRS \cite{Ferrara2014}. The available MAD techniques can be classified in two major types: (a) Single image based MAD (S-MAD) (b) Differential image based MAD (D-MAD). The Figure \ref{fig:MAD_techniques} shows the taxonomy of approaches in both MAD categories reported to date.

\subsection{Single Image Based MAD (S-MAD)}
\begin{figure}[htp]
	\centering
	\includegraphics[width=1\linewidth]{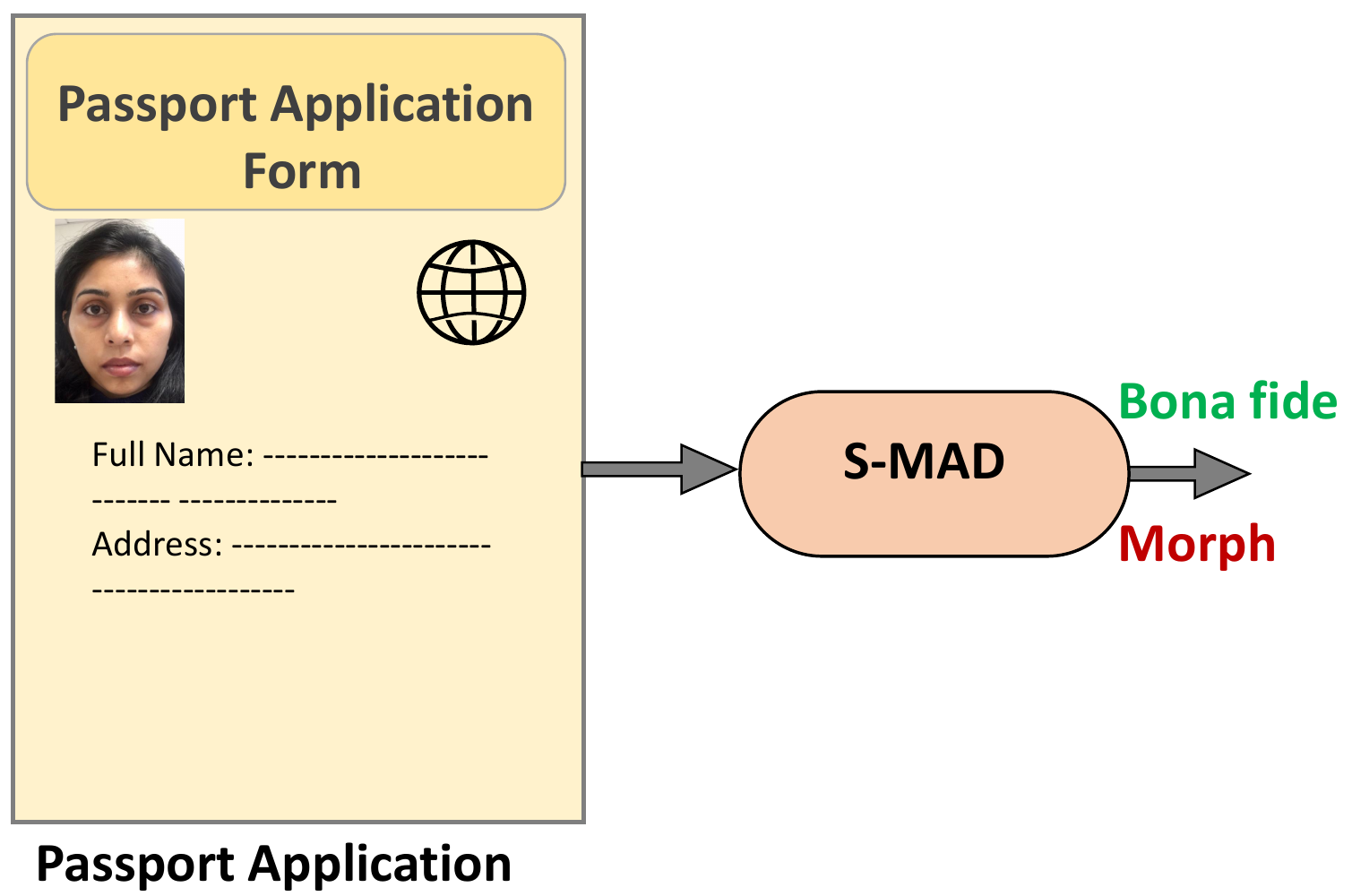}
	\vspace{-3mm}
	\caption{An example scenario illustrating the single image based morph attack detection in passport application scenario.}
	\label{fig:single_image_based_exa}
\end{figure}

The goal of S-MAD is to effectively detect a face morphing attack based on a single image presented to the algorithm. Figure \ref{fig:single_image_based_exa} illustrates the real life example for S-MAD in the passport application scenario, where a facial image is submitted by the applicant for biometric enrolment in the passport application process. This submitted image checked, in order to detect potentially a morph of the suspect image. The passport application can be initiated by the applicant either physically, when submitting his facial image through a web service \cite{nzvisa2019portal, ukpassport2019portal, india2019portal, netherlands2019portal}. Thus, depending on the type of the use case, the morphed image can be of two types: (a) Digital (b) Re-digitised (also commonly referred as print-scan). The S-MAD is challenging as it is expected to be robust to image quality variations, different type of sensors (cameras), different type of morph generation tools and different types of print-scan processes (e.g equipment and the parameter set chosen for the printing and scanning process.). 

\begin{table*}[htbp]
  \centering
  \caption{State-of-the-art on S-MAD}
  \resizebox{1.0\linewidth}{!}{
    \begin{tabular}{|p{11.665em}|p{4.365em}|p{13.6em}|p{25.265em}|p{4.965em}|}
    \hline
    \textbf{Reference} & \textbf{Detection} & \textbf{Approach} & \textbf{Algorithm} & \textbf{Database} \\
     & \textbf{Type} & & &  \bigstrut\\
    \hline
    \multirow{3}[2]{*}{Raghavendra et al. \cite{Raghavendra2016}} & \multirow{3}[2]{*}{S-MAD} & \multirow{3}[2]{*}{Texture based approach} & Local Binary Pattern (LBP)-SVM, Binary Statistical & \multirow{3}[2]{*}{Digital} \bigstrut[t]\\
    \multicolumn{1}{|r|}{} & \multicolumn{1}{r|}{} & \multicolumn{1}{r|}{} & Image Features (BSIF)-SVM, Image Gradient (IG)- & \multicolumn{1}{r|}{} \\
    \multicolumn{1}{|r|}{} & \multicolumn{1}{r|}{} & \multicolumn{1}{r|}{} & SVM   & \multicolumn{1}{r|}{} \bigstrut[b]\\
    \hline
    Makrushin et al. \cite{visapp17_Morph} & S-MAD & Quantized DCT coefficients & Benford features & Digital \bigstrut\\
    \hline
    \multirow{2}[2]{*}{Neubert et al. \cite{neubert2017Imagedegradation}} & \multirow{2}[2]{*}{S-MAD} & \multirow{2}[2]{*}{Image degradation approach} & \multirow{2}[2]{*}{Corner feature detector} & \multirow{2}[2]{*}{Digital} \bigstrut[t]\\
    \multicolumn{1}{|l|}{} & \multicolumn{1}{r|}{} & \multicolumn{1}{r|}{} & \multicolumn{1}{l|}{} & \multicolumn{1}{r|}{} \bigstrut[b]\\
    \hline
    Seibold et al. \cite{seibold2017ONLYVGGdetection} & S-MAD & Deep learning based approach & VGG19, Google Net, Alex Net & Digital \bigstrut\\
    \hline
    Raghavendra et al. \cite{raghavendra2017face} & S-MAD & Texture based approach & LBP, LPQ, BSIF, color textures & Print/Scan \bigstrut\\
    \hline
    Asaad et al. \cite{asaad2017topological} & S-MAD & Texture  based approach & Topological data analysis approach & Digital \bigstrut\\
    \hline
    Scherhag et al. \cite{Scher2017} & S-MAD & Texture and frequency based approach & LBP, LPQ, BSIF, 2DFFT with SVM classifier & Digital \newline Print/Scan \bigstrut\\
    \hline
    Kraetzer et al. \cite{Kraetzer:2017:MAP:3082031.3083244} & S-MAD & Texture based approach & Media forensics & Digital \bigstrut\\
    \hline
    Raghavendra et al. \cite{raghavendra2017transferable} & S-MAD & Deep CNN based approach & Feature fusion of fully connected layers of VGG19 and Alex Net & Digital \newline Print/Scan \bigstrut\\
    \hline
    Kraetzer et al. \cite{ PhotoID_MAD_Kraetzer_2017 } & S-MAD & Image life cycle model & Keypoints (SIFT, SURF, ORB, FAST, AGAST) and loss of edge operators (canny and sobel) & Digital \bigstrut\\
    \hline
    Hildebrandt et al. \cite{ StirTrace_Hildebrandt_2017} \cite{StirTrace_2018} & S-MAD & Stirtrace based approach & Multi-compression anomaly detection & Digital \bigstrut\\
    \hline
    Debiasi et al. \cite{PRNU} & S-MAD & Image degradation & Photo Response Non-Uniformity (PRNU) & Digital \bigstrut\\
    \hline
    Raghavendra et al. \cite{RagCVIP2018} & S-MAD & Steerable features & Luminance component extraction & Print/Scan \bigstrut\\
    \hline
    Hildebrandt et al. \cite{IWBF2017_StirTrace} & S-MAD & StirTrace & StirTrace face morph forgery detection & Print/Scan \bigstrut\\
    \hline
    Seibold et al. \cite{seibold2018reflection} & S-MAD & Image degradation & Specular reflection & Digital \bigstrut\\
    \hline
    Makrushin et al. \cite{BenfordLaw} & S-MAD & Quantized DCT coefficients & Benford features extracted from quantized DCT co-efficients & Digital \bigstrut\\
    \hline
    Neubert et al. \cite{Neubert2018ReducingTF} & S-MAD & Morph pipeline footprint detector & Benford features extracted from quantized DCT co-efficients & Digital \bigstrut\\
    \hline
    Spreeuwers et al. \cite{Luuk_FMdetect2018 } & S-MAD & Texture based approach & LBP-SVM, Down-up sampling & Digital \bigstrut\\
    \hline
    Scherhag et al. \cite{FeatureDifference} & S-MAD \newline D-MAD & Feature difference based approach & Pre-processing and feature extraction using texture descriptors , keypoint extractors, gradient estimators and deep learning based method & Digital \bigstrut\\
    \hline
    N Damer et al. \cite{naser_fusion_2019} & S-MAD & Multi detector fusion & LBPH, Transferable deep-CNN & Digital \bigstrut\\
    \hline
    Ferrara et al. \cite{Matteo_PrintScan_2019} & S-MAD & Deep learning & AlexNet, VGG19, VGG-Face16, VGG-Face2 & Print/Scan \bigstrut\\
    \hline
    Scherhag et al. \cite{Ulrich_fusion_2018} & S-MAD & Multi-algorithm fusion & Texture descriptors (LBP, BSIF), Keypoint extractors (SIFT, SURF), gradient estimators (HoG), Deep neural network & Digital \bigstrut\\
    \hline
    Debiasi et al \cite{PRNU_Debiasi_2018} & S-MAD &  PRNU & PRNU DFT magnitude histogram and PRNU DFT’s energy & Digital \bigstrut\\
    \hline
    Seibold et al. \cite{seibold2018accurate} & S-MAD & Complex multi-class pre-training & VGG-19 network & Digital \bigstrut\\
    \hline
    \multirow{2}[2]{*}{Venkatesh et al. \cite{DeepResidualNoise_IPTA_2019}} & \multirow{2}[2]{*}{S-MAD} & \multirow{2}[2]{*}{Color denoising based approach} & \multirow{2}[2]{*}{Denoising Deep Convolutional Neural Network} & \multirow{2}[2]{*}{Digital} \bigstrut[t]\\
    \multicolumn{1}{|r|}{} & \multicolumn{1}{r|}{} & \multicolumn{1}{r|}{} & \multicolumn{1}{r|}{} & \multicolumn{1}{r|}{} \bigstrut[b]\\
    \hline
    Scherhag et al. \cite{PRNU_TBIOM2019} & S-MAD & PRNU  & Spectral features and Spatial features & Print/Scan \bigstrut\\
    \hline
    Makrushin et al. \cite{Dempster_Shafer_2019} & S-MAD & Dempster-Shafer Theory & KeyPoints (SIFT,SUFT,FAST,ORB,AGAST, High Dim LBP, GoogleNet, VGG19 & Digital \bigstrut\\
    \hline
    Raghavendra et al. \cite{RagISBA2019} & S-MAD & Scale space approach & Color scale space features & Print/Scan \bigstrut\\
    \hline
    Neubert et al. \cite{eMRTD_MAD_Neubert_2019} & S-MAD & Frequency and Spatial domain feature space approach & Discrete Feature Transformation (DFT) , SURF, SIFT, ORB,  FAST, AGAST, Canny edge, SobelX, SobelY) & Digital \bigstrut\\
    \hline
    Seibold et al et al. \cite{StyleYourMorph_Ceibold_2019} & S-MAD & Style Transfer based approach & LBP, BSIF, Image degradation, Deep neural network (VGG19) & Digital \bigstrut\\
    \hline
    Venkatesh et al. \cite{Venkatesh_2020_WACV} & S-MAD & Color denoising based approach & Context Aggregation Network & Digital \bigstrut\\
    \hline
    Venkatesh et al. \cite{EnsembleFeatures_2020} & S-MAD & Ensemble of features based approach & LBP, HoG, BSIF & Print/Scan \bigstrut\\
    \hline
    \end{tabular}%
    }
  \label{tab:SMAD_techniques}%
\end{table*}%

As shown in the Figure \ref{fig:MAD_techniques} the existing S-MAD techniques can be further classified into five sub-types based on the features employed: (a) Texture features based S-MAD (b) Quality based S-MAD (c) Residual noise based S-MAD (d) Deep learning based S-MAD (e) Hybrid approaches for S-MAD. Table \ref{tab:SMAD_techniques} summarises the existing S-MAD techniques. In the following section, we briefly discuss the existing S-MAD techniques for the convenience of the reader. 
\paragraph{Texture Features Based S-MAD} The first work on using texture features was presented by Raghavendra et al. \cite{Raghavendra2016}. Following the initial work, several approaches were proposed as indicated in the Table \ref{tab:SMAD_techniques}. The popular texture based methods include Local Binary Patterns (LBP) \cite{Ojala-LBP-PatternRecognition-1996}, Local Phase Quantization (LPQ) features \cite{Ojansivu-LPQ-ICISP-2008} and Binarized Statistical Image Feature (BSIF) \cite{Kannala-BISF-ICPR-2012}. Further, these texture features were also extracted on different color channels \cite{raghavendra2017face} to obtain a robust detection performance. Variants of LBP and BSIF features together with Histogram of Oriented Gradients (HOG), Scale-Invariant Feature (SIFT) \cite{Lowe-ScaleInvariantFeatures-ICCV-1999} and Speeded-Up Robust Features (SURF) features \cite{Bay-SURF-CVIU-2008, Scher2017, Kraetzer:2017:MAP:3082031.3083244, Dempster_Shafer_2019, Ulrich_fusion_2018} is also widely explored in the reported works. The use of micro-texture based methods have indicated reasonable performance on both digital and print-scan type of S-MAD. While superior accuracy is reported on digital S-MAD with texture based features, the main limitation of these techniques is the generalizability across different image quality, imaging sensors and print-scan process \cite{Raja2020MorphingAD}. 

\paragraph{Quality Based S-MAD} The quality based techniques basically analyse the image quality features by quantifying the image degradation to identify the given image as morph or bona fide \cite{StirTrace_Hildebrandt_2017, IWBF2017_StirTrace, seibold2018reflection, PRNU_TBIOM2019, PRNU}. Several features such as detection double compression artefacts, Photo Response Non-Uniformity (PRNU), corner and edge distortions, reflection analysis and meta information from the images are commonly used to detect some distortion in a morphed image. Even though these techniques have indicated good performance on digital data, they have limited performance on print-scan data. However, the generalisation ability of these techniques are yet to be studied for different print and scan versions in the current literature \cite{Scherhag-PRNU-TBIOM-2019, Raja2020MorphingAD}.

\paragraph{Residual Noise Based S-MAD}
The residual noise based methods are designed to analyse the pixel discontinuity that may be largely impacted by the morphing process. The basic idea with this approach is to extract the noise patterns by subtracting the given image with de-noised version of the same image. The obtained noise patterns are further analysed to detect a morphing case. The first work in this direction was introduced in  \cite{DeepResidualNoise_IPTA_2019} based on CNN based de-noising on the colour channels. Further, the residual noise are effectively captured using deep CNN approach \cite{Venkatesh_2020_WACV}. The use of residual noise has indicated considerably good performance with generalisation capabilities across different digital datasets. However, these techniques are not evaluated on the print-scan face morphed datasets. 

\paragraph{Deep Learning Based S-MAD}
The success of deep learning approaches for image classification tasks has motivated researchers to embrace deep Convolutional Neural Networks (CNN) for face MAD. All existing works  are based on pre-trained networks and transfer learning. The first work in this direction is based on using the pre-trained networks such as AlexNet and VGG18 whose features are fused and classified to detect a morphing attack \cite{raghavendra2017transferable}. Following this, several deep CNN pre-trained networks such as AlexNet, VGG19, VGG-Face16, GoogleNet, ResNet18, ResNet150, ResNet50, VGG-Face2 and  Open face 
\cite{Matteo_PrintScan_2019}, \cite{naser_fusion_2019}, \cite{Venkatesh_2020_WACV}, \cite{StyleYourMorph_Ceibold_2019}, \cite{Ulrich_fusion_2018}, \cite{FeatureDifference}, \cite{seibold2017ONLYVGGdetection}, \cite{Dempster_Shafer_2019}, \cite{StyleYourMorph_Ceibold_2019} have been explored. Although, deep CNNs have indicated better performance when compared to hand-crafted texture descriptor based MAD methods on both digital and print-scan data, the generalisation capability of these approaches is limited across different print and scan datasets \cite{scherhag2020deep}. 

\paragraph{Hybrid S-MAD}
Hybrid approaches are based on more than one feature extractor or classifier that are combined to detect face morphing attacks. Several approaches are proposed that combine features, morphing detection scores or decision scores \cite{EnsembleFeatures_2020}, \cite{naser_fusion_2019}, \cite{Dempster_Shafer_2019}, \cite{Ulrich_fusion_2018}, \cite{RagCVIP2018}, \cite{RagISBA2019}. As these approaches combine more than one feature extraction and classifier, the MAD performance is generally superior when compared to single mode MAD techniques. Despite the superior performance, the computation cost is high and generalisation of the approach is not well established with respect to different types of print-scan processes.

A summary of advantages and demerits are illustrated in Table \ref{tab:AdvantageSMAD} for all different type of S-MAD techniques for the reference of the reader. 

\begin{table*}[htbp]
  \centering
   \caption{S-MAD Techniques: Advantages and Limitations}
    \resizebox{0.9\linewidth}{!}{
      \begin{tabular}{|p{8.085em}|p{17.835em}|p{17.165em}|}
    \hline
    \multicolumn{1}{|c|}{\textbf{Feature Type}} & \multicolumn{1}{c|}{\textbf{Advantages}}  & \multicolumn{1}{c|}{\textbf{Limitations}}  \bigstrut\\
    \hline
    Texture Features & - Easy to implement. \newline{} - Low computational costs. \newline{} -  Good performance when trained and tested with the same morph data types (digital/print-scan). \newline{} - Effective on digital morph face data &  - Lacks generalisation capabilities across both image resolution and morph data type (digital/print-scan). \newline{} -  Sensitive to image resolution. \newline{} -  Degraded performance with print-scan data. \bigstrut\\
    \hline 
    Image Quality Features & - Easy to implement. \newline{} -  Low computational cost. \newline{} -  Less sensitive to accurate segmentation of the face region. \newline{} - Can be used with different morph data types (digital/print-scan). & - Lacks generalisation across both image resolution and morph data types (digital/print-scan). \newline{} -  Sensitive to compressed data. \newline{} -  Not reasonable performance across different face morph data type (digital/print-scan). \bigstrut\\
    \hline
    Hybrid Features & - Good detection performance across different morph data types (digital/print-scan).
\newline{} - High detection performance when trained and tested with the same morph data type (digital/print-scan).
\newline{} -  Reasonable generalisability performance for different morph data types (digital/print-scan). & - Difficult to implement as it requires hyper parameter tuning. \newline{} - High computational costs. \newline{} - Requires optimisation of several hyper parameters.\bigstrut\\
    \hline
    Residual Noise Features & - Easy to implement.
\newline{} - Low computational costs. \newline{} - Highly accurate detection performance on digital morph data type. \newline{} - Less sensitive to face region. \newline{} - Generalisation ability across different image resolution. & - Applicable only for the digital morph data type. \newline{} - Promising results for high resolution images. \newline{} - Sensitive to image compression.\bigstrut\\
    \hline
    Deep CNN features & - Good performance when trained and tested with the same morph data type (digital/print-scan).
\newline{} - No need to train CNN from scratch as deep CNN indicate good detection performance. & - High computational cost. \newline{} -  Lacks generalisation across different face morph data types (digital/print-scan). \newline{} -  Training CNN from scratch requires large database. \bigstrut\\
    \hline
    \end{tabular}%
}
 \label{tab:AdvantageSMAD}%
\end{table*}%

\begin{table*}[htbp]
  \centering
  \caption{State-of-the-art on D-MAD}
  \resizebox{1.0\linewidth}{!}{
    \begin{tabular}{|p{9.9em}|p{4.5em}|p{7.565em}|p{21.835em}|p{5.1em}|}
    \hline
    \textbf{Reference} & \textbf{Detection} & \textbf{Approach} & \textbf{Algorithm} & \textbf{Database} \bigstrut\\
    & \textbf{Type} & & &  \bigstrut\\
    \hline
    M Ferrara et al. \cite{ferrara2018face} & D-MAD & Demorphing & Demorphing by image subtraction  & Print/Scan \bigstrut\\
    \hline
    M Ferrara et al. \cite{ferrara2018face} & D-MAD & Demorphing approach & Face verification & Digital \bigstrut\\
    \hline
    U Scherhag et al. \cite{Facial_LandMarks} & D-MAD & Landmark based approach & Distance based and Angle based feature extraction with Random Forest, SVM without kernel and SVM with radial basis function classifier & Digital \bigstrut\\
    \hline
    
    U Scherhag et al. \cite{FeatureDifference}  &  S-MAD \newline + D-MAD & Feature difference based approach & Pre-processing and feature extraction using texture descriptors, keypoint extractors, gradient estimators and deep learning based method & Digital \bigstrut\\

    \hline
    N Damer et al. \cite{naser_fusion_2019} & D-MAD & Multi detector fusion & LBPH, Transferable deep-CNN & Digital \bigstrut\\
    \hline
    \multirow{2}[2]{*}{J M Singh \cite{ Jag_ABC_gate_2019}} & \multirow{2}[2]{*}{D-MAD} & \multirow{2}[2]{*}{Deep learning} & \multirow{2}[2]{*}{SfS Net, Alexnet} & Digital \newline + Print/Scan \bigstrut[t]\\
    \multicolumn{1}{|r|}{} & \multicolumn{1}{r|}{} & \multicolumn{1}{r|}{} & \multicolumn{1}{r|}{} & \multicolumn{1}{r|}{} \bigstrut[b]\\
    \hline
    N Damer et al. \cite{MAD_LandmarkShift_Naser_2019} & D-MAD  & Landmark shift & Landmark detection, shift representation & Digital \bigstrut\\
    \hline
    F Peng et al. \cite{Peng_2019} & D-MAD & Face restoration by demorphing GAN & Symmetric dual network architecture & Digital \bigstrut\\
    \hline
    U Scherhag et al. \cite{scherhag2020deep} & D-MAD & Deep Face Representation & ArcFace Network, FaceNet algorithm & Digital \newline + Print/Scan \bigstrut[t]\\
    \hline
    \multirow{2}[2]{*}{C Seibold et al. \cite{SEIBOLD_2020}} & \multirow{2}[2]{*}{D-MAD} & \multirow{2}[2]{*}{Deep Learning} & \multirow{2}[2]{*}{Layer Wise Relevance Propagation (LRP)} & \multirow{2}[2]{*}{Digital} \bigstrut[t]\\
    \multicolumn{1}{|r|}{} & \multicolumn{1}{r|}{} & \multicolumn{1}{r|}{} & \multicolumn{1}{r|}{} & \multicolumn{1}{r|}{} \bigstrut[b]\\
    \hline
    D Ortego et al. \cite{BorderCtrl_MAD_2020} & D-MAD & Demorphing, Deep CNN based  & Auto-encoders & Digital \newline + Print/Scan \bigstrut[t]\\
    \hline
    \end{tabular}%
    }
  \label{tab:DMAD}%
\end{table*}%


\begin{table*}[htbp]
  \centering
  \caption{D-MAD Techniques: advantages and limitations}
   \resizebox{0.8\linewidth}{!}{
    \begin{tabular}{|l|p{17.335em}|p{18.265em}|}
    \hline
    \multicolumn{1}{|p{9.935em}|}{\textbf{Algorithm type}} & \textbf{Advantages} &  \textbf{Limitations} \bigstrut\\
    \hline
    \multicolumn{1}{|l|}{\multirow{3}[2]{*}{Feature difference}} & -  Easy to implement. & - High computational costs. \bigstrut[t]\\
          & - Reasonable detection performance across varying image quality and resolutions. & - Detection performance is sensitive to the type of image data and features. \\
          & \multicolumn{1}{r|}{} & - Detection performance is sensitive to the segmentation of face region. \bigstrut[b]\\
    \hline
    \multicolumn{1}{|l|}{\multirow{4}[2]{*}{Demorphing}} & - Easy to implement. &  - Performance is sensitive to the facial poses and imaging conditions. \bigstrut[t]\\
          & - Moderate computational time. & - Requires constrained image data. \\
          & - High detection accuracy with constrained conditions. & - Fails with facial poses and lighting variations.  \\
          & - Can visualise the de-morphed face if the suspected image is morphed & - Prior knowledge on blending factor (or alpha factor) is required.   \bigstrut[b]\\
    \hline
    \end{tabular}%
    }
  \label{tab:DMADAdvan}%
\end{table*}%

\begin{figure}[htp]
	\centering
	\includegraphics[width=1\linewidth]{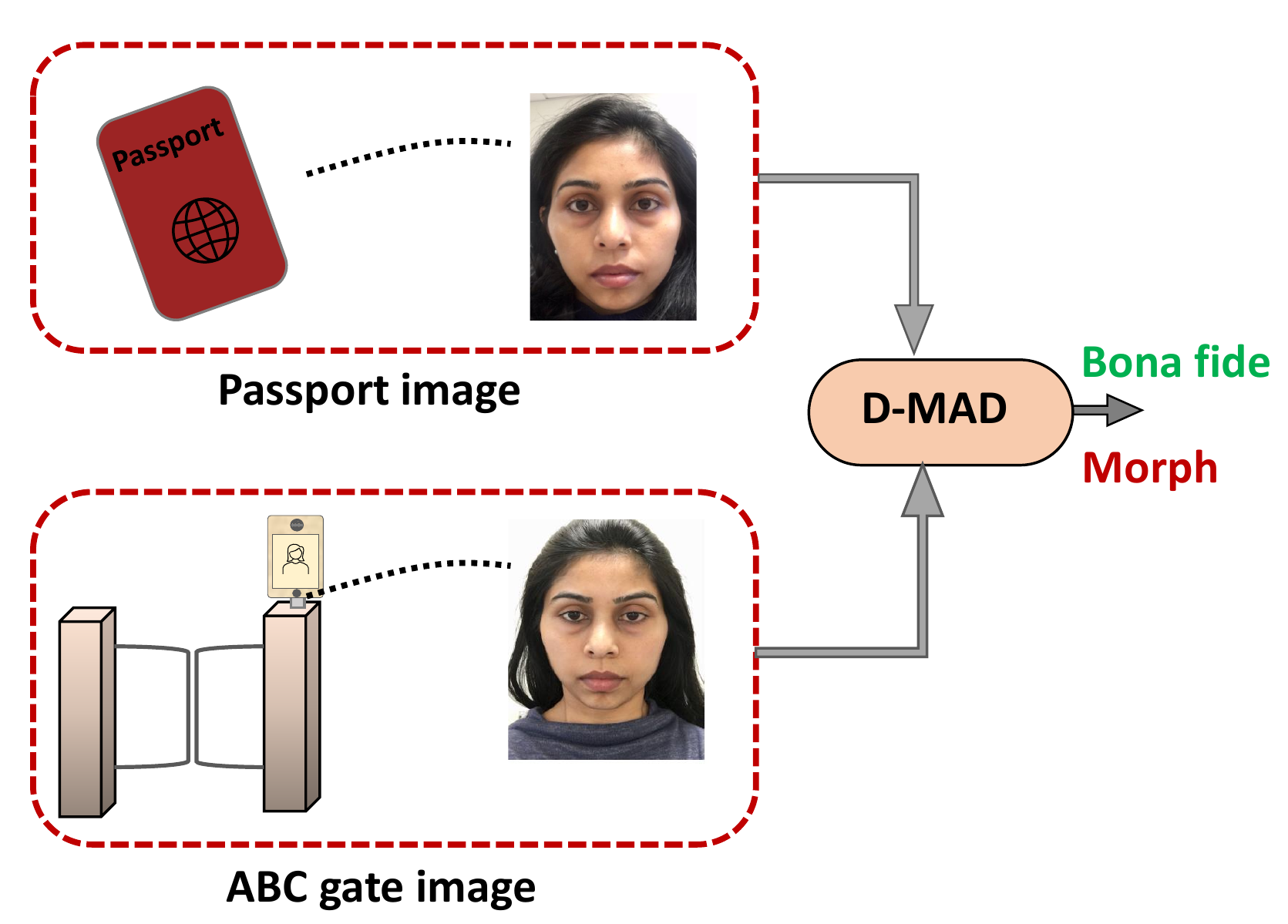}
	\vspace{-3mm}
	\caption{An example scenario illustrating the differential image based morph attack detection in passport control scenario}
	\label{fig:Ref_image_based_exa2}
\end{figure}
\subsection{Differential Image Based MAD (D-MAD)}

The objective of Differential Image Based (D-MAD) approaches is to make a decision on the suspected image as morph or bona fide when a corresponding image captured in a trusted environment is available. The D-MAD technique suits well for the border crossing scenario, where the suspected morph image can be obtained from the passport that can then be compared against the live captured face image (or trusted image) from the Automated Border Control (ABC) gates \cite{BorderCtrl_MAD_2020}. Figure \ref{fig:Ref_image_based_exa2} illustrates the application of D-MAD, especially in a border control scenario. The taxonomy of D-MAD techniques is presented in Figure \ref{fig:MAD_techniques} and can be divided broadly into two types: (a) Feature difference based D-MAD (b) Demorphing. Table \ref{tab:DMAD} summarizes the existing D-MAD techniques that are briefly discussed below:
\paragraph{Feature Difference Based D-MAD} 
The basic idea of this approach is to subtract the features computed on both the suspected morph image and a live image captured in trusted environment. The features are further classified by computing the difference in the feature vectors to detect the morphing attack. To this extent, several feature extraction techniques are studied that includes texture information, 3D information, gradient information, landmark points and deep feature information \cite{scherhag2020deep}, \cite{Jag_ABC_gate_2019}, \cite{naser_fusion_2019}, \cite{Facial_LandMarks}, \cite{MAD_LandmarkShift_Naser_2019}.  Based on the reported results, the deep CNN features have indicated the best performance \cite{scherhag2020deep}. The majority of the existing works are reported for the use case with digital images except the recent work where a print and scan dataset is explored with improved results \cite{scherhag2020deep, NISTReport2020}. 

\paragraph{Demorphing}
The face demorphing techniques inverts the morphing procedure and reveals the component images that are used to generate the morphed image. The first work in this direction was proposed by Ferrara et al. \cite{ferrara2018face} that was designed to work with the landmark based morph generation. Recent works in this direction are based on using deep CNNs \cite{BorderCtrl_MAD_2020} \cite{Peng_2019}. These techniques are robust when image quality is good, however the detection performance degrades when a face image is captured in real-life conditions with pose and lighting variations that are commonly encountered in ABC gates. Table \ref{tab:DMADAdvan} further presents the advantages and limitations of existing D-MAD techniques.

\section{Performance Metrics}
\label{sec:performancemetrics}

In this section, we discuss the performance evaluation metrics that are widely used in the literature and the publicly available competitions to benchmark the performance of MAD techniques.

\subsection{Vulnerability Assessment of FRS } 
For a morph image to be deemed as a significant threat to FRS, it is necessary to establish the threat potential. Most of the works provide the threat potential by measuring the vulnerability of FRS. We therefore provide a brief overview of suitable metrics for establishing the relevance of morph attacks through vulnerability metrics. The goal of face vulnerability analysis is to measure if the generated morphed face image can be verified against all contributory data subjects. Thus, when a morphed face image is enrolled into a FRS and probed with another image from a contributing subjects, the FRS must successfully verify all contributory subjects corresponding to the pre-set verification threshold. In the most of the work, the threshold of the FRS has been adjusted to correspond to false match rate (FMR) of $0.1\%$ following the guidelines from FRONTEX \cite{frontex2015best}.

\begin{figure}
	\centering
	\includegraphics[width=0.95\linewidth]{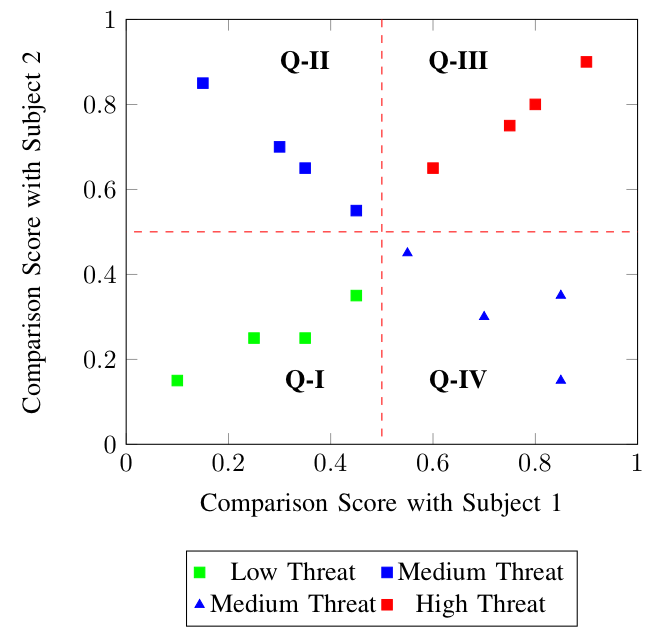}
	   \caption{Threats of morphed images with respect to comparison score against both contributing subjects. The figure illustrates the morphed images crossing the threshold of 0.5 (i.e., laying in quadrant Q-III) are effective attacks with severe threat to FRS as compared to Q-II and Q-IV. 
	    }
   \label{fig:vulnerability-example}
\end{figure}

Figure \ref{fig:vulnerability-example} illustrates the example of the vulnerability plots that represent the scatter of comparison scores from FRS. The sample vulnerability plot is simplified for visualization purposes to provide an illustration of the vulnerability analysis. Figure \ref{fig:vulnerability-example} can be interpreted using four different quadrants. The first quadrant (bottom left quadrant) $Q-I$ indicates the morphed image being not verified to neither of the two contributing data subjects. Thus, a large number of comparison scores in the first quadrant indicates that the morph generation method is not strong enough to deceive the COTS FRS (or, in other words, the morphing image is not a severe threat). The second quadrant (top left quadrant) $Q-II$ indicates that the morphed image can be verified to data subject-2 (one of the contributing subjects) only. Therefore the morphed images poses a medium threat. The third quadrant (top right quadrant) $Q-III$ indicates that the morphed image being verified to both contributing data subjects (subject-1 and subject-2). Thus, the larger the number of comparison scores in this quadrant, the higher is the threat and the vulnerability of the analysed FRS with respect to morphed images. The fourth quadrant (bottom right quadrant) $Q-IV$ indicates the a morphed image can be verified as data subject-1 only. Therefore the morphed images poses again a medium threat to the FRS. 
  
To mathematically quantify the vulnerability of a FRS towards morphed face images the following metrics are developed and adapted in the literature.
\paragraph{\textbf{Mated Morph Presentation Match Rate (MMPMR)}} This metric was initially proposed by Scherhag et al. in \cite{Scherhag-MorphingAttacks-MorphingTechniques-BIOSIG-2017}. It defines the proportion of morphed images verified with its contributing images. 
 \begin{equation}
 MMPMR = \frac{1}{M} \sum_{m=1} ^{M}  {[min_{n=1 \ldots N_{m}^{}}  S_{m} ^{n} ] > \tau }
 \label{Eqa:MMPMR}
\end{equation}

Where, $M$ is the number of morphed images, $N_{m}^{}$ is the total number of subjects contributing for morph $m$. $ S_{m} ^{n}$ is the comparison score for mated morph for the morph $m$ of $n^{th}$  subject and $\tau$ being the threshold of FRS at a chosen FAR. 

The rational of the MMPMR is that a morphing attack succeeds if all contributing subjects are verified successfully against the morph image. While MMPMR  considers multiple comparisons, which are related to multiple authentication attempts. That may not always be the case. 
A successor of the MMPMR metric named Fully Mated Morph Presentation Match Rate (FMMPMR) was introduced by Venkatesh et al., \cite{Venkatesh_2020_IJCB} to address the quadrants employed for vulnerability assessment as shown in Figure \ref{fig:vulnerability-example}. The details of the FMMPMR are provided below.

\paragraph{\textbf{Fully Mated Morph Presentation Match Rate (FMMPMR)}} This metric defines the proportion of morphed images verified with its contributing subjects again under the condition that the morphed image verifies successfully against both contributing subjects \cite{Venkatesh_2020_IJCB}. This metric takes further into account the number of successful attempts of probe images from the contributing subjects towards the morph image. 

\begin{dmath}
	FMMPMR = \frac{1}{P} \sum_{M,P}^{} {(S1_{M}^{P} > \tau) AND (S2_{M}^{P} > \tau) \\ \ldots AND  (Sk_{M}^{P} > \tau)}
	\label{Eqa:FMMPMR}
\end{dmath}
Where $P = {1, 2, \ldots, p}$ represent the number of attempts made by comparing all probe images from the contributing subject against $M^{th}$ morphed image,  $K = {1, 2, \ldots, k}$ represents the number of contributing data subjects to the  constitution of the generated morphed image (in our case  $K=2$), $Sk_{M}^{P}$ represents the comparison score of the $K^{th}$ contributing subject obtained with $P^{th}$ attempt (in this case the $P^{th}$ probe image from the dataset) corresponding to $M^{th}$ morph image and $\tau$ represents the threshold value corresponding to FMR = 0.1$\%$. The FMMPMR metric verifies the morphed image with its contributing subjects and in addition it takes into account the number of attempts. It is therefore a relevant and realistic metric to quantify the vulnerability and establishing the true attack strength of a morph generation method.

\subsection{MAD Performance Metrics} 
The robustness of MAD algorithms is measured using the performance metrics defined in the International Standard ISO/IEC 30107-3 \cite{ISO-IEC-30107-3-PAD-metrics-170227} and are applicable to report the morphing attack detection performance. Since MAD performance can be visualised as a binary classification problem, the following metrics are widely used to benchmark the MAD algorithms.  
    \begin{itemize}
        \item  \textbf{Attack Presentation Classification Match Rate (APCER):} Defines the proportion of attack samples incorrectly classified as bona fide face images.   
        \item  \textbf{Bona fide Presentation Classification Match Rate (BPCER):} Defines the proportion of bona fide images incorrectly classified as attack samples. 
    \end{itemize}


\begin{figure}[htp]
	\centering
	\resizebox{0.49\textwidth}{!}{
		\includegraphics[width=0.8\linewidth]{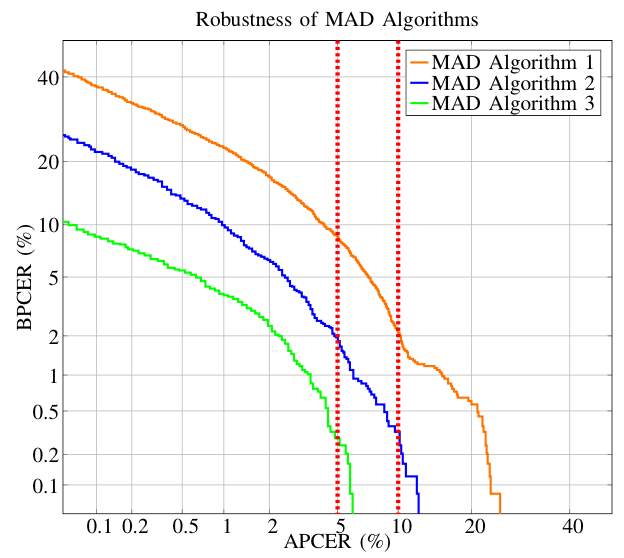}}
	\caption{Sample illustration of the detection accuracy of a MAD algorithms at different operating points with a detection error trade-off curve (DET). As noted from the figure, the MAD Algorithm 3 performs best at chosen APCER of $5\%$ or/and $10\%$}
	\label{fig:mad-strength}
\end{figure}

However, it is not possible to optimise both APCER and BPCER jointly, it is thus natural to set (or fix) either BPCER or APCER and report the result with a dependency of the other metric (either APCER/BPCER). Most of the works have reported the results by setting a predefined security level (e.g. indicating the maximum proportion of morph accepts they wan to tolerate)  and then fixing APCER accordingly to values of $@1\%$, $5\%$ or/and $10\%$ \cite{raghavendra2017transferable, NISTReport2020, Raja2020MorphingAD}. As shown in Figure~\ref{fig:mad-strength}, the MAD Algorithm 3 would be preferred at a given APCER of $5\%$ or/and $10\%$ in the benchmark with the two other algorithms. 

\subsection{Joint Evaluation of MAD and Vulnerability} 

In a real-life scenario, the FRS may operate with MAD sub-system in an integrated processing. For a succesful attack it is therefore important that the morphed face image can invade the enrolment process and also will reach a match to probe images from the contributory subjects. To quantify the performance of the vulnerability in the presence of MAD a metric called Actual Mated Morph Presentation Match Rate (AMPMR) was recently proposed in  \cite{QinLe_PartialMorph} that can be written as follows: 

\begin{dmath}
	AMPMR(th_{fa}, th_{mad} ) = \frac{1}{N} \sum_{i=1}^{N} {(((\min_{j=1,..M{i}}  S C_{ij})  > th_{fa}) AND (SC_{mad-i}> th_{mad})) }
	\label{Eqa:AMPMR}
\end{dmath}

Where the total number of morph images is denoted by N. $SC_{ij}$ is the face recognition score of the $i^{th}$ morphed image when compared to a probe sample of the $j^{th}$ contributor.  $M{i}$ is the number of contributors to the morphed image. $SC_{mad-i}$ is the MAD score of the $i^{th}$ sample. Based on this metrics, higher values of the AMPMR indicate a severe vulnerability. 

\section{Public Evaluation and Benchmarking}
\label{sec:Publiccompetitions}
In this section, we summarize face morphing evaluations that publicly benchmarking morphing attack detection performance. At the time of this writing there two such benchmarks: The Face Recognition Vendor Test (FRVT) Part 4: MORPH - Performance of Automated Face Morph Detection \cite{MeiNGAN-morph-FRVT-2020}, Bologna-SOTAMD: Evaluation of Differential Morph Attack Detection and Single Image Morph Attack Detection \cite{Raja2020MorphingAD}. These benchmarks have provided a common platform that includes datasets, evaluation protocols and the computational environment. The platforms provides a trustworthy assessment of algorithms that can be obtained for submitted algorithms. In the following, we briefly describe the databases used in each platform and the performance achieved by various algorithms that are presented. 

\subsection{NIST-FRVT Part 4: MORPH - Performance of Automated Face Morph Detection}
The FRVT MORPH test was opened in June 2018 to provide a common platform for independent testing of face MAD technologies and to ensure a common assessment methodology. The dataset involved in the evaluation is created using different morphing methods, which have the objective to reflect low-quality morphing (generated using freely available tools), automated morph generation (generated using an automatic tool without any human intervention) and high-quality morph generation (that are generated with commercial morphing software and additional post-processing that is carried out to mask potential artifacts). The evaluation is carried out for both S-MAD and D-MAD techniques. However, the probe data used in the D-MAD evaluation is not effectively stemming from  Automatic Border Gates (ABC). Several algorithms are evaluated and the majority of the participants in the competition to date are from academic institutes. Most of the submissions for S-MAD are based on texture features, while for the D-MAD, both face de-morphing and differential features based techniques are evaluated. Based on the recent evaluation report, it can be noted that:
\begin{enumerate}
    \item None of the algorithms has indicated a reliable detection performance meeting the FRONTEX operational requirement \cite{frontex2015best} and thus face morphing attack detection remains a challenging task.
    \item The quality of the morph generation has direct impact on the performance of both S-MAD and D-MAD techniques.
\end{enumerate}
In the S-MAD category, the use of hybrid features \cite{RagISBA2019} has indicated a better performance over other MAD methods, while among the methods in D-MAD category the approach of latent feature differences based on ArcFace features \cite{scherhag2020deep} reaches the best detection performance.

\subsection{Bologna-SOTAMD: Differential Morph Attack Detection}
The Bologna-SOTAMD benchmark was opened for evaluation in 2019 and provided a common evaluation platform to benchmark D-MAD techniques. The  Bologna-SOTAMD D-MAD benchmark consists of the database collected in the European SOTAMD project \cite{SOTAMD_project} using real Automatic Border Gates (ABC). The morphing is carried out using both automated approaches with open source and commercial software. Several MAD techniques are benchmarked that include both face de-morphing and feature difference methods, the details of evaluation protocol and performance of various submitted algorithms can be found in \cite{Raja2020MorphingAD}. Among the multiple algorithms evaluated, it can be noted that the existing D-MAD techniques are not robust enough to detect the face morphing attacks in accordance to FRONTEX operational requirement \cite{frontex2015best} highlighting the challenge of MAD again. The use of the feature difference based D-MAD technique indicates the better performance over face de-morphing techniques.  The best result with D-EER = 3.36 $\%$ is reported on digital  and D-EER = 3.36 $\%$ is reported on print-scan data. 	

\subsection{Bologna-SOTAMD: Single Morph Attack Detection}
The Bologna server is also hosting a public benchmark for S-MAD since 2020. The S-MAD dataset is constructed using high-quality passport images similar to the one used in the real passport. The morphed images are generated using both commercial (FantaMorph, FaceFusion) and open source (Triangulation with facial-landmarks) face morphing software.  Post-processing is carried out using automatic and manual processes to overcome the artefacts generated using the face morphing software. For more information on the database and evaluation protocol, readers can refer to  \cite{Raja2020MorphingAD}. 
As the evaluation started only recently, not many algorithms are benchmarked yet in on the Bologna S-MAD platform. The baseline performance reported a D-EER of 37.10\% and 38.99\% on print-scan and digital morphed images. These preferences measure indicates the challenges in detecting face morphing images using S-MAD techniques.  

\subsection{Discussion on Public Evaluation}
Thus, based on the above discussion on the publicly available benchmarks and competitions, it can be noted that the reliable detection of the face morphing attacks remains challenging. It can be pointed out that the S-MAD technique's performance is severely degraded compared to that of the D-MAD methods. This can be attributed to the availability of additional information (availability of another image) that can be used to make the final decision. The interesting outcome of these competitions indicates that the use of hybrid features based S-MAD technique has shown improved generalisability across various morph generation methods. At the same time, the feature difference method used in the context of D-MAD has indicated a more robust performance in both benchmarks.  

\begin{figure}[htp]
	\centering
	\includegraphics[width=1\linewidth]{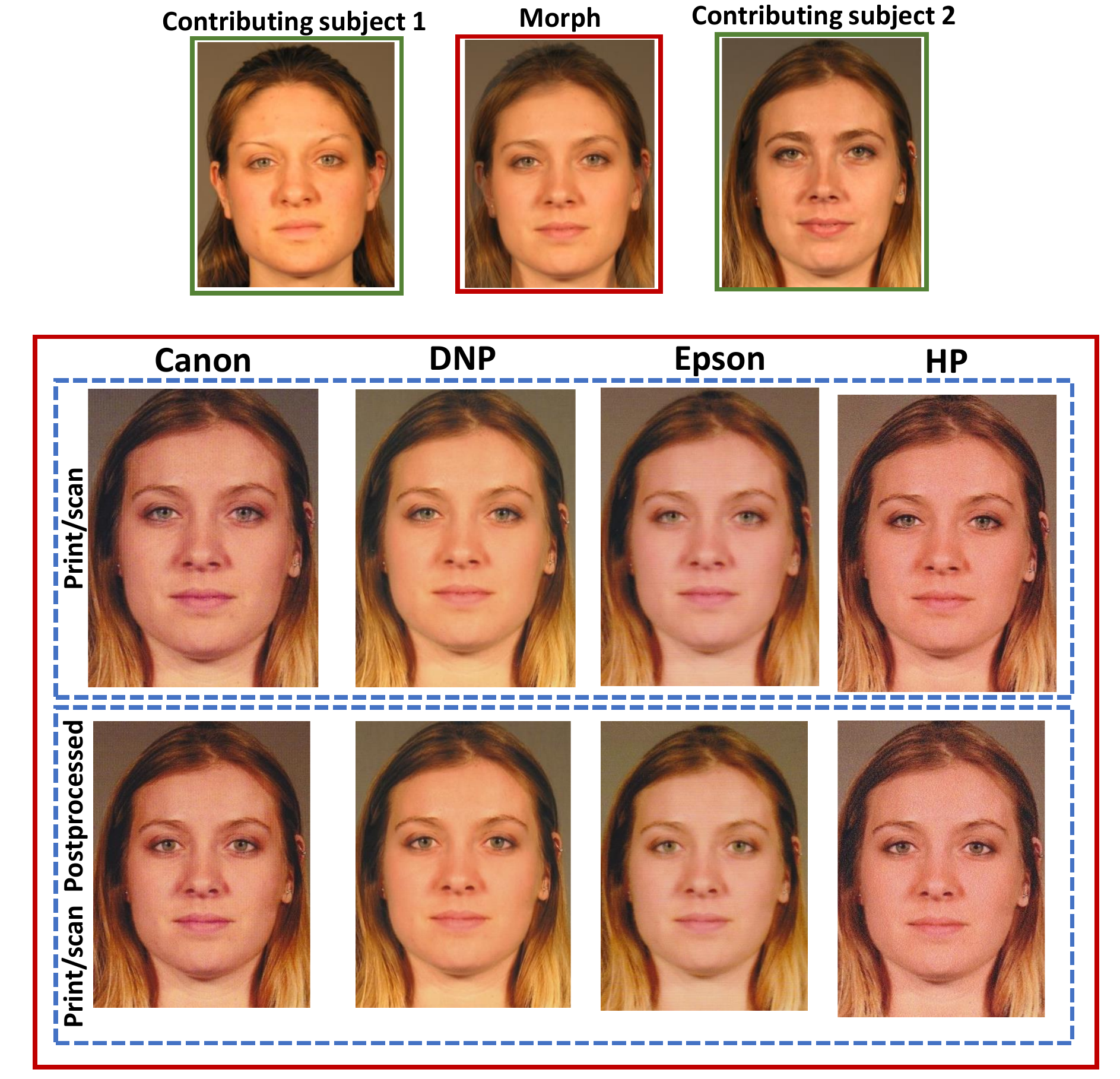}
	\vspace{-3mm}
	\caption{Example of  print/scan images and post-processed images. The variation of the data quality across different printers and scanners can be noted which challenge the MAD algorithms.}
	\label{fig:Morph_PSImages}
\end{figure}

\section{Open Challenges}
\label{sec:challenges}

The research topic of face morphing and detection has received greater interest from both research and governmental stakeholders. This has resulted in intensive research activities in terms of studying the vulnerability of COTS FRS and the development of several MAD techniques to reliably detect said attacks. However, there are still several challenges and open issues that need to be addressed. In the following section, we present such challenges and open issues in the field of face morphing attack detection. 

\paragraph{\textbf{Unavailability of large-scale public datasets with variation}}
The unavailability of large scale face morphing datasets reflecting the real-life scenario hinders the development of robust MAD. Further, considering the different modes of morph attacks (digital and print-scan), it is necessary to generate and evaluate the MAD algorithms on digital and print-scan datasets. However, the generation of large print/scan databases is quite expensive and tedious.  Additionally, the databases cannot be shared publicly due to licensing restrictions as well as to privacy and GDPR \cite{GDPR_EU} concerns. Hence there is a limitation to access the existing morphing databases. Even though the publicly available benchmarks are now hosting large-scale databases, those datasets can only test the submitted MAD algorithms. They can not serve the further development of MAD algorithms. However, the systematic generation of morphed face images with various types of morphing software combined with different types of print-scan processes must result in large-scale databases that become available for researchers in order to achieve significant progress in MAD. 
 
\paragraph{\textbf{Generalisability of MAD techniques}}
The generalisation of the MAD techniques is crucial in achieving reliable performance in real-life border control scenarios. However, the existing MAD techniques are evaluated only on known types of face morph generation technique and the known source of re-digitisation (printer and scanner type) except in NIST-FRVT Part 4: MORPH - Performance of Automated Face Morph Detection and Bologna-SOTAMD. Figure \ref{fig:Morph_PSImages} illustrates the variation in the morphed image quality due to different types of printer-scanner type.  The performance reported in benchmarking study \cite{MeiNGAN-morph-FRVT-2020}, \cite{FVCOngoingMorphingEval}, \cite{Raja2020MorphingAD} also indicated the degraded performance of MAD techniques on both D-MAD and S-MAD when tested on unknown source of generation. More significant degradation is noted with S-MAD methods attributed to the learning-based systems that can learn the decision policy based on the known data. These factors further limit the applicability of learning-based MAD techniques if they are not trained on the large scale dataset with all real-life variants. Thus, it is essential to device the MAD approach robust in detecting face morphing attacks. 

\paragraph{\textbf{Selection of data subjects for morphing}}
It can also be noted that in earlier studies, the morphing images are generated by randomly selecting the contributing data subjects. It is well established assumption that the morphing attack will be more successful with both human observers and machine (FRS), if the candidate data subjects that are selected based on a look alike measure.  Some of the recent works  \cite{Doppelhenger_IJCB_2020},  \cite{raghavendra2017face},  \cite{RightFaces_Naser_2019}, \cite{Venkatesh_2020_IJCB} describe the selection of data subjects in the morphing process. However, the systematic study of these existing methods towards the impact on the FRS vulnerability, the detection performance of both human observers and automatic MAD detection methods still needs to be investigated. 

\paragraph{\textbf{Variation with face co-variates}}
The critical aspect that is not systematically studied with MAD is the role of face co-variates that includes aging, gender, ethnicity, identification factor, image post-processing and image quality.  The preliminary study on the aging effect on morphing vulnerability and detection is presented in \cite{Venkatesh_2020_IJCB} that has revealed the influence of aging for face morphing vulnerability. The variation of face co-variates is more influential on S-MAD techniques, while for D-MAD techniques, the imaging quality plays a vital role. As the images are captured using ABC gate in D-MAD, the influence of varying illumination due to day and night light settings need to be investigated. Additionally, the live captured images at the ABC gate may be acquired with eyeglasses or hair occlusions which are not investigated yet. Thus, it is essential to benchmark both D-MAD and S-MAD techniques in a real-life scenario with all those co-variates. Another aspect that was not investigated yet with its impact on MAD is potential face beautification. It is expected that the face images are beautified prior to applying for the passport in many countries \cite{rathgeb2020differential}. As the beautification process changes the image properties, it is essential to understand both vulnerability and MAD for this particular problem. 

\paragraph{\textbf{Performance metrics}}
Considering that the face morphing attack detection is emerging as the new operational problem, there is only a slow convergence towards harmonized testing and reporting. The publicly available benchmarking and competitions have employed the ISO/IEC metrics \cite{ISO/IEC2015a} to benchmark the detection performance of MAD techniques. However, there are no standardized metrics yet, to evaluate the vulnerability of FRS with respect to morphing attacks. Therefore, there is a strong need for a standardised vulnerability evaluation metric incorporating the experience from both practitioners and researchers working on face MAD. The availability of an international standard using ISO/IEC, together with commonly used vulnerability metrics, are discussed in Section \ref{sec:performancemetrics} which needs further efforts. 

\paragraph{\textbf{Component based morphing}}
Almost all literature have studied the face morphing as a holistic problem with full face image morphing. Le el al., \cite{QinLe_PartialMorph} introduced partial face morphing that includes a preliminary study on morphing only specific regions in the face region. Extensive experiments reported indicate that a partial morphing with eye and nose poses a severe threat to commercial face recognition systems \cite{QinLe_PartialMorph}. However, the systematic evaluation on high quality face images are yet to be studied together with the impact on human expert observers (for example border guards and super recognizer). 

\paragraph{\textbf{Identical Twins and Look-Alike}}
Influence of morphing on identical twins and look-alike is an interesting problem that needs a systematic study within the scope of morphing. The vulnerability of FRS due to face morphing image generated from identical twins and look-alike subjects need to be studied on the large-scale databases. 

\paragraph{\textbf{User convenience}}
The design of user-convenient (or user-friendly) MAD systems plays a crucial role in making detection sub-systems deployable in real-time applications. Thus, there is a need to design face MAD systems that allow minimum user intervention (both from operators and applicants). This fact needs to be considered when designing D-MAD techniques that are tailored for ABC systems. 


\section{Conclusion}
\label{conclusion}
Face recognition systems have gained high amount of trust for security related applications. However, morphing attacks on face recognition systems can be a hindrance to establish a secure society. Further there have been various morphing attack detection techniques being proposed by several researchers to effectively detect morphed images. However, improvements in the deep learning and machine learning techniques have resulted in generation of high quality morphs using various new techniques. Hence, generalising a morph attack detection is still predicted to be a long way ahead considering the basic challenge of obtaining large public databases with variation and different morph generation techniques. 
In this paper, we have detailed the advancement of different types of morph generation techniques. Alongside a brief overview of the different types of morphing attack detection techniques is reported together with the corresponding performance metrics. We have also provided a brief discussion on the challenges faced in this field to develop a robust technique to detect morphs to serve as reference for future works. 

\section{Acknowledgement}
\label{acknowledge}

The authors would like to thank European Commission for supporting this work funded by iMARS project. The content of this work represents the views of the authors only and is their sole responsibility. The European Commission does not accept any responsibility for use that may be made of the information it contains.











{
	\bibliographystyle{ieee}
	\bibliography{FaceMorph}
}


\end{document}